\documentclass[12pt]{article}
\usepackage{array}
\usepackage{algorithm}
\usepackage{algorithmic}
\usepackage{amsthm}
\usepackage{amsmath}
\usepackage{amsfonts}
\usepackage{amssymb}

\usepackage{color}

\usepackage[english]{babel}
\usepackage{booktabs}
\usepackage{caption}
\usepackage{enumitem} 
\usepackage{float}
\usepackage{graphicx}
\usepackage [colorlinks, linkcolor=blue, anchorcolor=blue, citecolor=blue, urlcolor=blue] {hyperref}
\usepackage[utf8]{inputenc}
\usepackage{longtable}
\usepackage{makecell}
\usepackage{multirow}
\usepackage[figuresright]{rotating}
\usepackage{textcomp}
\usepackage{url}
\usepackage{verbatim}
\usepackage{cite}
\usepackage[caption=false,font=normalsize,labelfont=sf,textfont=sf]{subfig}

\newtheorem{definition}{Definition}[section]
\newtheorem{pf}{Proof}
\newtheorem{thm}{Theorem}
 
\newtheorem{lemma}{Lemma}
\newtheorem{example}{Example}
\newtheorem{remark}{Remark}
\usepackage[dvipsnames]{xcolor}
\usepackage{bm}
\hypersetup{colorlinks,urlcolor=blue}
\usepackage{svg}
\usepackage{adjustbox}

\usepackage[numbers,sort&compress]{natbib}
\title{Decision-oriented joint optimization of evidence fusion based on event-conditioned credibility\footnote{This work has been submitted to the Elsevier for possible publication. Copyright may be transferred without notice, after which this version may no longer be accessible.}}
\author{Chaoxiong Ma \thanks{chaoxiongma@mail.nwpu.edu.cn} \and Yan Liang \thanks{liangyan@nwpu.edu.cn} \and Huixia ZHANG \thanks{zhanghuixia@mail.nwpu.edu.cn} \and Hao Sun \thanks{sun\_hao233@mail.nwpu.edu.cn}}

\newcolumntype{R}[2]{%
    >{\adjustbox{angle=#1,lap=\width-(#2)}\bgroup}%
    l%
    <{\egroup}%
}

\date{} 

\begin{document}

\def\spacingset#1{\renewcommand{\baselinestretch}%
{#1}\small\normalsize} \spacingset{1}

\maketitle
In decision-level fusion tasks involving heterogeneous sources with unequal precision and potential anomalies, evidence deviating from the majority may be either critical evidence supporting the correct decision or anomalous evidence supporting an incorrect event. Existing credible evidence fusion (CEF) methods primarily assess credibility through inter-evidence comparisons and may consequently underestimate critical evidence. This paper proposes event-conditioned credibility to characterize the relative credibility of evidence under different candidate-event hypotheses. A decision-oriented joint optimization model then couples credibility calculation, evidence fusion, and event decision through candidate-event probabilities. The model is expressed as a continuous self-mapping on the probability simplex. A direct fixed-point iteration provides the default fast solver, while a Kuhn simplicial search supplies a mesh-dependent approximate fixed point if the direct iteration forms a periodic orbit. A plausibility--belief arithmetic--geometric divergence is further proposed to calculate event-conditioned credibility. In the labeled numerical examples, the proposed method gives credibility rankings that better reflect the contribution of evidence to the ground truth and provides greater support for the correct event than representative CEF methods. Monte Carlo tests additionally demonstrate a high empirical fixed-point attainment rate over the tested settings.


\section{Introduction}
\label{sec:introduction}

In the multisource information fusion and decision-making system, information sources with different observation perspectives and feature spaces form local judgments based on observations and domain knowledge, which are integrated at the decision level to determine the unknown state or event. Evidence reasoning (ER) \cite{LI2024120462,06_geng2021earc} converts heterogeneous information into mass functions on a common frame of discernment (FoD) and combines them for decision making. It has been widely applied in fault diagnosis \cite{02_xiao_2021_anoval,XU2024119995}, target recognition \cite{09_geng2020multi,10_song2018sensor}, service evaluation \cite{04_fei2020extended}, signal classification \cite{ZHANG2023119107,08_han2022trusted}, and human activity recognition \cite{qiu2022multi}. However, multisource information may contain both high quality information critical to decision-making and anomalous information caused by outliers, faults, or interference. If represented as evidence, such information may provide strong support for mutually exclusive candidate events, resulting in a high degree of conflict among the evidences. Applying Dempster’s combination rule (DCR)~\cite{11_dempster2008upper} to highly conflicting evidence may produce a counterintuitive result\cite{12_wang2021new,zhao2022survey}.

To do so, two categories of improvements have been developed: rule modification and credible evidence fusion (CEF). Rule-modification methods modify the combination rule, as in the approaches proposed by Smets\cite{13_smets1990combination}, Dubois\cite{14_dubois1988representation}, and Yager\cite{15_yager1987dempster}. However, they often lose the associative law and/or commutative law of DCR \cite{xiao2022gejs}, and hence fail to deal with the high conflict caused by sensor failures \cite{16_xiao2019multi}. These losses also lead to two drawbacks: the fusion output depends on the combination order, making the search for an optimal order NP-hard; moreover, the extensibility of probability is destroyed, losing statistical interpretability \cite{17_haenni2002alternatives}.

The CEF treats DCR as a natural extension of probability theory and attributes counterintuitive fusion results to disturbed evidence rather than to the rule itself \cite{18_huang2021cross}. Accordingly, it preprocesses evidence using credibility before fusion. Murphy\cite{19_murphy2000combining} first assigns equal credibility to all evidence, while Deng\cite{20_yong2004combining} later introduces the evidence difference measure (EDM) for differentiated credibility according to the sources' varying abilities to provide accurate information. Subsequent studies have extensively explored EDM design, such as evaluating evidence availability using entropy \cite{deng2016deng}, managing conflict using correlation coefficients \cite{jiang2019new}, and refined EDM functions (EDMFs) for improved rationality \cite{16_xiao2019multi,21_liu2011combination,22_xiao2020new}. Despite these differences, existing CEF methods generally follow a one-shot, four-step sequential procedure: constructing the EDM matrix (EDMM), calculating evidence credibility based on the EDMM, preprocessing evidence using credibility, and fusing the preprocessed evidence by DCR. This line of research has established an effective framework for conflict management and has been successfully applied in various domains, including multisensor smart home systems \cite{li2024inconsistency}, fault diagnosis with deception identification \cite{cui2023bgc}, and EEG analysis based on Rényi divergence \cite{zhu2022ageneralized}, thereby laying a solid foundation for credible information fusion.

The above CEF methods calculate credibility through inter-evidence comparisons, assigning higher credibility to evidence with smaller EDMs to the majority and treating larger EDMs as indications of possible anomalies. This premise is reasonable if evidence differences mainly arise from anomalies or measurement errors. Differences in source precision, observation perspective, feature discriminability, and domain knowledge may also enable a few sources to provide critical evidence that supports the ground truth more strongly than the majority. Critical and anomalous evidence may both have large EDMs to the majority, although they play opposite roles in identifying the ground truth. Inter-evidence comparisons assign low credibility in both cases, leaving critical evidence underutilized in decision-level fusion.

Whether evidence is critical or anomalous depends on the ground truth, which is itself the unknown event to be determined through ER. To address this issue, this paper proposes event-conditioned credibility, which characterizes the relative credibility of evidence under different event hypotheses by treating each candidate event in turn as the ground truth. The overall credibility of each evidence is obtained by aggregating its event-conditioned credibility according to the candidate-event probabilities, while these probabilities are determined by the fusion result of the credibility-weighted evidences. Credibility calculation, evidence fusion, and event decision are thereby formulated as an interdependent joint optimization problem.

The main contributions of this paper are summarized as follows.
\begin{enumerate}[label=\arabic*)]
	\item To address the potential underestimation of critical evidence by credibility calculations based on inter-evidence comparisons, event-conditioned credibility is proposed. It evaluates the relative credibility of evidence under different event hypotheses by treating each candidate event in turn as the ground truth, thereby establishing a relationship between evidence credibility and candidate events.
	\item A decision-oriented joint optimization fusion model (JOFM) for ER is developed. The model determines evidence credibility by aggregating event-conditioned credibility according to candidate-event probabilities, while updating these probabilities using the fusion result of credibility-weighted evidences. Credibility calculation, evidence fusion, and event decision are thus formulated as an interdependent joint problem.
	\item A plausibility--belief arithmetic--geometric divergence (PBAGD) is proposed to measure the difference between evidence and candidate-event evidence and thereby calculate event-conditioned credibility. By jointly using the lower and upper support represented by belief and plausibility functions, the divergence accounts for support conveyed by both singleton and compound focal elements. It satisfies symmetry, nonnegativity, and nondegeneracy.
\end{enumerate}

The remainder is organized as follows. Section \ref{sec: Problem formulation} formulates the problem. Section \ref{sec:The conditional credibility} introduces event-conditioned credibility and PBAGD. Section \ref{sec:Iterative credible evidence fusion} presents the JOFM. Section \ref{sec:Experiment and application} reports experiments. Section \ref{sec: conclusion} concludes.

\section{Problem formulation}
\label{sec: Problem formulation}
Consider decision-level multisource evidence fusion in which exactly one of $n$ mutually exclusive and collectively exhaustive events in the FoD $\Omega = \{\tilde{A}_j\}_{j=1}^{n}$ occurs as the ground truth. Each of the $N$ information sources forms a local judgment based on its acquired information and domain knowledge. To represent this judgment and its uncertainty, the $i$th source provides a mass function $\boldsymbol{m}_i\!=\![m_i(A_1), \cdots\!, m_i(A_r)]^T$, where $A_k$ is the $k$th nonempty proposition of the power set $2^{\Omega}=\{A|A\subseteq\Omega\}$, $r\!=\!2^n\!-\!1$, and $T$ denotes transpose. The mapping $m_i: 2^\Omega \to [0,1]$ satisfies $\sum_{A_k\subseteq\Omega} m_i(A_k) = 1$ and $m_i(\emptyset)=0$.

The mass functions provided by different sources may differ substantially because their information quality and event discrimination capabilities are not identical. For example, in target recognition, infrared sensors are more sensitive to high-temperature targets, millimeter-wave radars better capture the electromagnetic scattering characteristics of metallic targets, and low-frequency ultra-wideband radars obtain stronger echoes from plastic targets. Differences in observation perspective, feature discriminability, and domain knowledge may further make a source particularly informative for certain events. Evidence critical to decision making is therefore event dependent and may differ markedly from other evidences. Large evidence differences may also arise from outliers, faults, interference, deception, or erroneous judgments. In the presence of both, critical and anomalous evidence may strongly support mutually exclusive basic events, resulting in high conflict.

\subsection{A general process of the existing CEF}
A common method in CEF is the expected evidence method, which first calculates a credibility-weighted average evidence as
\begin{equation}
\label{eq:average_evidence} 
\boldsymbol{m}_{avg} = \sum\nolimits_{i=1}^N {Cred_i \boldsymbol{m}_i}
\end{equation}
where credibility $Cred_i$ represents the relative likelihood that ``$\boldsymbol{m}_i$ is the most credible.''
\begin{equation}
\mathrm{Cred}_i\geq 0,
\qquad
\sum\nolimits_{i=1}^{N}\mathrm{Cred}_i=1.
\end{equation}
The average evidence is then combined with itself $N-1$ times to obtain fusion output $\boldsymbol{m}_{\oplus}$
\begin{equation}
\label{eq:repeated_combination}
\boldsymbol{m}_{\oplus} = \underbrace {\boldsymbol{m}_{avg} \oplus\boldsymbol{m}_{avg}\oplus\boldsymbol{m}_{avg}}_{N}
\end{equation}
where $\oplus$ is the DCR operator\cite{shafer1976mathematical}. For $\boldsymbol{m}_i$ and $\boldsymbol{m}_j$, DCR is defined as $\forall A_k \in 2^\Omega$ and $B,C\subseteq\Omega$
\begin{equation}
\left( {{m_i} \oplus {m_j}} \right)\left( A_k \right) \!\!= \!\!
\begin{cases}
0 \!\!\!&, \text{if}\ A_k = \emptyset \\
\frac{{\sum\nolimits_{B \cap C = A_k} {{m_i}\left( B \right){m_j}\left( C \right)} }}{{1 - {\sum _{B \cap C = \emptyset }}{m_i}\left( B \right){m_j}\left( C \right)}} \!\!\!&,{\text{otherwise.}}
\end{cases}
\label{eq:dempster_rule}
\end{equation}

Existing CEF methods derive $Cred_i$ from the EDMM $\mathbf{D}=[d_{ih}]\in\mathbb{R}^{N\times N}$ where $d_{ih}$ is the EDM between $\boldsymbol{m}_i$ and $\boldsymbol{m}_h$, including average-support-based calculation \cite{24_zhang2022novel}
\begin{equation}
\label{eq:average_support_credibility}
Cred_i = \frac{\sum\nolimits_{h = 1,h \ne i}^N 1/d_{ih}}{\sum\nolimits_{j = 1}^N \sum\nolimits_{h = 1,h \ne j}^N 1/d_{jh}}
\end{equation}
and those based on the principal eigenvector \cite{21_liu2011combination,23_sun2019new}:
\begin{equation}
\mathbf{G}=[g_{ih}],\quad g_{ih}=\eta(d_{ih}),\quad
\mathbf{G}\boldsymbol{v}=\lambda_{\max}\boldsymbol{v},
\qquad
Cred_i = \frac{v_i}{\sum\nolimits_{j = 1}^N v_j},
\label{eq:eigenvector_credibility}
\end{equation}
where $\eta(\cdot)$ is a positive, non-increasing function, and $\lambda_{\max}$ and $\boldsymbol{v}$ are the principal eigenvalue and corresponding eigenvector of $\mathbf{G}$, respectively.
\subsection{Underestimation of Critical Evidence}
In this subsection, a multisensor evidence fusion example from \cite{28_wang2021new} is used to illustrate how credibility based solely on inter-evidence EDMs may underestimate critical evidence.
\begin{example}
	\label{ex1}
	(\textbf{Multisensors fault diagnosis}) Consider three distinct automotive faults: low oil pressure $\tilde{A}_1$, intake system leakage $\tilde{A}_2$, and electromagnetic valve jamming $\tilde {A}_3$. The state of vehicle is set to ``low oil pressure''. As shown in Table \ref{tab1:Example1Evidence}, five evidences are reported individually, with Sensor 1--4 working normally and Sensor 5 reporting an anomaly.
\end{example}
\begin{table}[!h]
	\setlength\tabcolsep{1.5mm
		\centering
		\renewcommand{\arraystretch}{1}
		\caption{Multisensors evidence report~\cite{28_wang2021new}.}
		\label{tab1:Example1Evidence}
		\renewcommand{\arraystretch}{1}
		\begin{tabular}{cccccc|cccccc}
			\toprule [1.25pt]
			&$\boldsymbol{m}_1$ &$\boldsymbol{m}_2$  &$\boldsymbol{m}_3$  &$\boldsymbol{m}_4$ &$\boldsymbol{m}_5$&&$\boldsymbol{m}_1$ &$\boldsymbol{m}_2$  &$\boldsymbol{m}_3$  &$\boldsymbol{m}_4$ &$\boldsymbol{m}_5$ \\
			\midrule [0.5pt]
			$\{\tilde A_1\}$ 	 & 0.70 &0.70  &0.65  &\textbf{0.75} &0.00  &
			$\{\tilde A_2\}$	 & 0.10 &0.00  &0.15  &\textbf{0.00} &0.20 \\
			$\{\tilde A_3\}$	 & 0.00 &0.00  &0.00  &\textbf{0.05} &0.80&
			$\Omega=\{\tilde{A}_i\}_{i=1}^{3}$	 &0.20  &0.30  &0.20  & \textbf{0.20}&0.00 \\
			\bottomrule [1.25pt]
	\end{tabular}}
\end{table}
Evidences $\boldsymbol{m}_1$--$\boldsymbol{m}_4$ primarily support the
ground truth $\tilde{A}_1$. Among them, $\boldsymbol{m}_4$ provides the
strongest support and is therefore regarded as critical evidence. In
contrast, the anomalous evidence $\boldsymbol{m}_5$ strongly supports the
incorrect event $\tilde{A}_3$.

\begin{table}[!h]
	\setlength\tabcolsep{1.5mm
		\centering
		\renewcommand{\arraystretch}{1}
		\caption{Evidence credibility based on four EDMFs. }
		\label{tab2:Example1Credibility}
		\begin{tabular}{ccccccccc}
			\toprule [1.25pt]
			\multirow{2}{*}{Evidence} &	\multicolumn{2}{c}{Dismp\cite{21_liu2011combination}} &\multicolumn{2}{c}{RB\cite{22_xiao2020new}} &\multicolumn{2}{c}{BJS\cite{16_xiao2019multi}} &\multicolumn{2}{c}{PBLBJS\cite{28_wang2021new}} \\ \cline{2-9}
			&Credibility&Rank&Credibility&Rank&Credibility&Rank&Credibility&Rank\\
			\midrule[0.5pt]
			$\boldsymbol{m}_1$ &\textbf{0.2476}&1&\textbf{0.2393}&1&\textbf{0.2471}&1&0.2426&3\\
			$\boldsymbol{m}_2$ & 0.2475        &2&0.2246        &3& 0.2106         &4&0.2567&2\\
			$\boldsymbol{m}_3$ & 0.2411        &4&0.2261        &2& 0.2428&2&\textbf{0.2755}&1\\
			$\boldsymbol{m}_4$ & 0.2432        &3&0.2126        &4& 0.2300         &3&0.2013&4\\
			$\boldsymbol{m}_5$ & 0.0206        &5&0.0973        &5& 0.0695         &5&0.0239&5\\
			\bottomrule [1.25pt]
	\end{tabular}}
\end{table}

\begin{table}[!h]
	\setlength\tabcolsep{0.8mm}   
	\caption{Fusion results under different methods.\label{tab3:Example1Fusion}}
	\centering
	\renewcommand{\arraystretch}{1}
	\begin{tabular}{ccccc|ccccc}   
		\toprule[1.25pt]
		& $\{\tilde A_1\}$ & $\{\tilde A_2\}$ & $\{\tilde A_3\}$ & $\Omega$ & & $\{\tilde A_1\}$ & $\{\tilde A_2\}$ & $\{\tilde A_3\}$ & $\Omega$ \\
		\midrule[0.5pt]
		Dismp\cite{21_liu2011combination} & 0.9833 & 0.0082 & 0.0032 & 0.0053 &
		RB\cite{22_xiao2020new} & 0.9914 & 0.0034 & 0.0043 & 0.0008 \\
		BJS\cite{16_xiao2019multi} & 0.9937 & 0.0030 & 0.0025 & 0.0008 &
		PBLBJS\cite{28_wang2021new} & 0.9957 & 0.0026 & 0.0008 & 0.0009 \\	
		\bottomrule[1.25pt]
	\end{tabular}
\end{table}
Table~\ref{tab2:Example1Credibility} shows that all four methods, Dismp\cite{21_liu2011combination}, BJS\cite{16_xiao2019multi}, RB\cite{22_xiao2020new}, and belief divergence measure based on belief and plausibility function (PBLBJS)\cite{28_wang2021new}, assign the lowest credibility to $\boldsymbol{m}_5$, but none ranks the critical evidence $\boldsymbol{m}_4$ first. Instead, higher credibility is assigned to evidences that are more consistent with the other normal evidences. Nevertheless, all methods correctly identify the ground truth $\tilde{A}_1$ under the maximum Pignistic probability decision rule (MPPDR)~\cite{31_xiao2021ceqd}, with fused support above $0.98$, as shown in Table~\ref{tab3:Example1Fusion}. This indicates that inter-evidence comparison effectively suppress $\boldsymbol{m}_5$, but the critical evidence is underestimated.

\begin{table}[!h]
	\setlength\tabcolsep{1.5mm}   
	\centering
	\renewcommand{\arraystretch}{1}
	\caption{Multisensors evidence report.}
	\label{tab4:Example1Evidence_modify}
	\renewcommand{\arraystretch}{0.9}
	\begin{tabular}{cccccc|cccccc}   
		\toprule[1.25pt]
		& $\boldsymbol{m}_1$ & $\boldsymbol{m}_2$ & $\boldsymbol{m}_3$ & $\boldsymbol{m}_4$ & $\boldsymbol{m}_5$ &
		& $\boldsymbol{m}_1$ & $\boldsymbol{m}_2$ & $\boldsymbol{m}_3$ & $\boldsymbol{m}_4$ & $\boldsymbol{m}_5$ \\
		\midrule[0.5pt]
		$\{\tilde A_1\}$ & 0.35 & 0.40 & 0.35 & \textbf{0.70} & 0.00 &
		$\{\tilde A_2\}$ & 0.30 & 0.10 & 0.25 & \textbf{0.10} & 0.30 \\
		$\{\tilde A_3\}$ & 0.00 & 0.20 & 0.20 & \textbf{0.00} & 0.70 &
		$\Omega=\{\tilde{A}_i\}_{i=1}^{3}$ & 0.35 & 0.30 & 0.20 & \textbf{0.20} & 0.00 \\
		\bottomrule[1.25pt]
	\end{tabular}
\end{table}
\begin{table}[!h]
	\setlength\tabcolsep{0.25mm
		\centering
		\renewcommand{\arraystretch}{1.2}
		\caption{Evidence credibility based on four EDMFs. }
		\label{tab5:Example1Credibility_modify}
		\begin{tabular}{ccccccccc}
			\toprule [1.25pt]
			\multirow{2}{*}{Evidence} &	\multicolumn{2}{c}{Dismp\cite{21_liu2011combination}} &\multicolumn{2}{c}{RB\cite{22_xiao2020new}} &\multicolumn{2}{c}{BJS\cite{16_xiao2019multi}} &\multicolumn{2}{c}{PBLBJS\cite{28_wang2021new}} \\ \cline{2-9}
			&Credibility&Rank&Credibility&Rank&Credibility&Rank&Credibility&Rank\\
			\midrule[0.5pt]
			$\boldsymbol{m}_1$ &0.2241&3&0.2107&3&0.1960&3&0.2574&3\\
			$\boldsymbol{m}_2$ & 0.2423        &2&0.2341        &2& 0.2566         &2&\textbf{0.3426}&1\\
			$\boldsymbol{m}_3$ & \textbf{0.2544}        &1&\textbf{0.2426}        &1& \textbf{0.2953}&1&0.3146&2\\
			$\boldsymbol{m}_4$ & 0.1743        &4&0.1889        &4& 0.1667         &4&0.0699&4\\
			$\boldsymbol{m}_5$ & 0.1050        &5&0.1237        &5& 0.0854         &5&0.0154&5\\
			\bottomrule [1.25pt]
	\end{tabular}}
\end{table}
\begin{table}[!h]
	\setlength\tabcolsep{0.8mm}   
	\caption{Fusion results under different methods.\label{tab6:Example1Fusion_modify}}
	\centering
	\renewcommand{\arraystretch}{1}
	\begin{tabular}{ccccc|ccccc}   
		\toprule[1.25pt]
		& $\{\tilde A_1\}$ & $\{\tilde A_2\}$ & $\{\tilde A_3\}$ & $\Omega$ &
		& $\{\tilde A_1\}$ & $\{\tilde A_2\}$ & $\{\tilde A_3\}$ & $\Omega$ \\
		\midrule[0.5pt]
		Dismp\cite{21_liu2011combination} & 0.6608 & 0.1835 & 0.1225 & 0.0332 &
		RB\cite{22_xiao2020new} & 0.7664 & 0.1290 & 0.0989 & 0.0057 \\
		BJS\cite{16_xiao2019multi} & 0.7852 & 0.1240 & 0.0847 & 0.0061 &
		PBLBJS\cite{28_wang2021new} & 0.7805 & 0.1423 & 0.0679 & 0.0093 \\
		\bottomrule[1.25pt]
	\end{tabular}
\end{table}
To further examine how the underestimation of critical evidence affects fusion results, $\boldsymbol{m}_1$--$\boldsymbol{m}_3$ are replaced by the more uncertain evidences in Table~\ref{tab4:Example1Evidence_modify}, while the $\boldsymbol{m}_4$ and $\boldsymbol{m}_5$ are still critical and anomalous evidences. As shown in Table~\ref{tab5:Example1Credibility_modify}, all four methods still rank $\boldsymbol{m}_5$ last but rank the critical evidence $\boldsymbol{m}_4$ fourth because it is less consistent with $\boldsymbol{m}_1$--$\boldsymbol{m}_3$. All methods still select $\tilde{A}_1$, but the fused support decreases to $0.6608$--$0.7852$, as reported in Table~\ref{tab6:Example1Fusion_modify}. Notably, Dismp produces a support of $0.6608$, which is lower than the $0.70$ originally provided by $\boldsymbol{m}_4$. Assigning low credibility to critical evidence may therefore leave its valid information underutilized, even if the final decision remains correct.

These examples show that inter-evidence comparisons effectively suppress evident anomalies but may also underestimate critical evidence that deviates from the majority. Deviation from the majority alone is therefore insufficient to determine evidence credibility. Distinguishing critical evidence from anomalous evidence requires considering their respective relationships to the
true event. However, the true event is unknown and must be inferred from the fusion result, which itself depends on evidence credibility. The core problem addressed in this paper is therefore to couple credibility calculation, evidence fusion, and event decision to jointly determine evidence credibility, the fusion result, and candidate-event probabilities.

\section{Event-conditioned credibility}
\label{sec:The conditional credibility}
\subsection{Event-conditioned credibility based on event evidence}
\label{subsec:The proposal of conditional credibility}
Let $\Phi$ be a categorical variable indexing the relative credibility assigned to the $N$ evidences, and let $\phi_i=\{\Phi=i\}$ and $Cred_i=p(\phi_i)$. Here, $p(\phi_i)$ denotes the normalized relative credibility of $\boldsymbol{m}_i$ rather than the probability of a physical event. Let $p(\tilde{A}_j)$ denote the probability that candidate event $\tilde{A}_j$ is the ground truth. Since the candidate events are mutually exclusive and collectively exhaustive, the law of total probability gives
\begin{equation}
Cred_i=p(\phi_i)=\sum\nolimits_{j=1}^n{p(\phi_i|{\tilde{A}}_j)p({\tilde{A}}_j)}
\label{eq:total_probability_credibility}
\end{equation}

\begin{definition}[Event-conditioned credibility]
	\label{definition:conditional credibility}
	$p(\phi_i\mid\tilde{A}_j)$ is termed the event-conditioned credibility of evidence $\boldsymbol{m}_i$, representing its relative credibility under the hypothesis that $\tilde{A}_j$ is the ground truth. It satisfies $p(\phi_i|\tilde{A}_j)\geq 0$ and $\sum\nolimits_{i=1}^{N}p(\phi_i\mid\tilde{A}_j)=1$.
\end{definition}
If $\tilde{A}_j$ is the ground truth, evidence providing stronger support for $\tilde{A}_j$ should be assigned higher relative credibility. Quantifying the event-conditioned credibility	therefore requires measuring the proximity of evidence $\boldsymbol{m}_i$ to $\tilde{A}_j$. To represent the candidate event in the same evidence space as $\boldsymbol{m}_i$, the event evidence associated with $\tilde{A}_j$ is defined as
\begin{equation}
m_{\tilde{A}_j}\left(B\right) = 
\begin{cases}
1 &, \text{if}\ B = \{\tilde{A}_j\} \\
0 &,{\text{otherwise.}}
\end{cases}
\text{,   }\quad\forall{B}\subseteq\Omega.
\label{eq:event_evidence}
\end{equation}
The proximity of evidence $\boldsymbol{m}_i$ to candidate event	$\tilde{A}_j$ is then quantified by the EDM between $\boldsymbol{m}_i$ and $\boldsymbol{m}_{\tilde{A}_j}$, denoted by $\tilde{d}_{ji} = d(\boldsymbol{m}_{\tilde{A}_j},\boldsymbol{m}_i)$, where $d(\cdot,\cdot)$ is an EDMF. Let $\kappa(\cdot)$ be a positive and nonincreasing kernel. Event-conditioned credibility is calculated by normalizing kernel values
\begin{equation}
p(\phi_i|{\tilde{A}}_j) = \frac{\kappa(\tilde{d}_{ji})}{\sum\nolimits_{l=1}^{N}\kappa(\tilde{d}_{jl})},
\qquad \kappa(d)>0,
\label{eq:ecc_general_kernel}
\end{equation}

Two kernel specifications used in this study are the inverse-distance and exponential kernel
\begin{equation}
\kappa_{\mathrm{inv}}(d)=\frac{1}{d},
\qquad d>0,
\label{eq:inverse_kernel}
\end{equation}
\begin{equation}
\kappa_{\mathrm{exp}}(d;\tau)=\exp(-\tau d),
\qquad \tau>0.
\label{eq:exponential_kernel}
\end{equation}
For $\tilde d_{ji}>0$, if the distance coefficient $\tau_{ji}={\ln\tilde d_{ji}}/{\tilde d_{ji}}$, there is $\tilde 1/d_{ji} =\exp(-\tau_{ji}\tilde d_{ji})$.

\subsection{Plausibility--belief arithmetic--geometric divergence}
\label{subsec: PBAGD}
To calculate event-conditioned credibility, PBAGD is defined to measure the EDM between evidence and event evidence.

For evidence $\boldsymbol{m}_i$, a normalized mapping is constructed from its belief function $\mathrm{Bel}_{\boldsymbol{m}_i}$ and plausibility function $\mathrm{Pl}_{\boldsymbol{m}_i}$\cite{05_jiao2016combining}:
\begin{equation}
PB_{\boldsymbol{m}_i}(A_k)=\frac{\exp\!\left(\mathrm{Bel}_{\boldsymbol{m}_i}(A_k)\right)+\exp\!\left(\mathrm{Pl}_{\boldsymbol{m}_i}(A_k)\right)
}{\displaystyle\sum\nolimits_{B\subseteq\Omega}\left[\exp\!\left(\mathrm{Bel}_{\boldsymbol{m}_i}(B)\right) + \exp\!\left(\mathrm{Pl}_{\boldsymbol{m}_i}(B)\right)\right]
}, \qquad{A_k}\subseteq\Omega,
\label{eq:pb_mapping}
\end{equation}
where
\begin{equation}
\mathrm{Bel}_{\boldsymbol{m}_i}(A_k)=\sum\nolimits_{B\subseteq A_k}m_i(B),\quad \mathrm{Pl}_{\boldsymbol{m}_i}(A_k)=\sum\nolimits_{B\cap A_k\neq\emptyset}m_i(B).
\label{eq:belief_plausibility_function}
\end{equation}

For any two evidences $\boldsymbol{m}_i$ and $\boldsymbol{m}_h$, PBAGD is defined as
\begin{equation}
\mathrm{PBAGD}(\boldsymbol{m}_i,\boldsymbol{m}_h)={}\sum_{A_k\subseteq\Omega}
\frac{PB_{\boldsymbol{m}_i}(A_k)+PB_{\boldsymbol{m}_h}(A_k)}{2}\log\frac{PB_{\boldsymbol{m}_i}(A_k)+PB_{\boldsymbol{m}_h}(A_k)}{2\sqrt{PB_{\boldsymbol{m}_i}(A_k)PB_{\boldsymbol{m}_h}(A_k)}}.
\label{eq:pbagd}
\end{equation}
Since the exponential mapping yields $PB_{\boldsymbol{m}}(A_k)>0$ for every nonempty $A_k\subseteq\Omega$, PBAGD is well defined for normalized mass functions on a finite FoD.

\begin{lemma}
	\label{Lemma 1}
	For normalized mass functions on $\Omega$, Eq.(\ref{eq:pb_mapping}) is injective if $PB_{\boldsymbol{m}_i}(A)$ is evaluated over $2^\Omega$. That is, if $PB_{\boldsymbol{m}_1}(A)=PB_{\boldsymbol{m}_2}(A), \forall A\subseteq\Omega$, then $\boldsymbol{m}_1=\boldsymbol{m}_2$.	
\end{lemma}
\begin{pf}
	Let $q_{\boldsymbol m}(A)=\exp(Bel_{\boldsymbol m}(A))+\exp(Pl_{\boldsymbol m}(A))$ and $Z_{\boldsymbol m}=\sum_{B\subseteq\Omega}q_{\boldsymbol m}(B)$. Equality of the two PB mappings at $A=\Omega$, where $q_{\boldsymbol m}(\Omega)=2e$, implies equality of their normalization factors and hence $q_{\boldsymbol m_1}(A)=q_{\boldsymbol m_2}(A)$ for every $A\subseteq\Omega$. For a nontrivial $A$, set $x=\exp(Bel_{\boldsymbol m}(A))$ and $y=\exp(Bel_{\boldsymbol m}(\bar A))$. Then $q_{\boldsymbol m}(A)/q_{\boldsymbol m}(\bar A)=x/y$, while $q_{\boldsymbol m}(A)q_{\boldsymbol m}(\bar A)=s+2e+e^2/s$, where $s=xy\in[1,e]$. Since the latter expression is strictly decreasing on $[1,e]$, the values of $q_{\boldsymbol m}(A)$ and $q_{\boldsymbol m}(\bar A)$ uniquely determine $x$ and $y$. Thus, the two belief functions are identical on $2^\Omega$, and Möbius inversion gives $\boldsymbol m_1=\boldsymbol m_2$.
\end{pf}
\begin{thm}
	For any two normalized evidences $\boldsymbol{m}_1$ and $\boldsymbol{m}_2$ defined on the same finite FoD, the PBAGD satisfies the following properties\textup{:}
	\begin{enumerate}[label=\arabic*)]
		\setlength{\leftmargin}{-1em}
		\setlength{\labelsep}{0em}
		\setlength{\itemindent}{-1.5em}
		\item Symmetry\textup{:} $\mathrm{PBAGD}(\boldsymbol{m}_1,\boldsymbol{m}_2)\!=\! \mathrm{PBAGD}(\boldsymbol{m}_2,\boldsymbol{m}_1)$;
		\item Nonnegativity\textup{:} $\mathrm{PBAGD}(\boldsymbol{m}_1,\boldsymbol{m}_2)\!\geq\!0$;
		\item Nondegeneracy\textup{:} $\mathrm{PBAGD}(\boldsymbol{m}_1,\boldsymbol{m}_2)\!=\!0 \Leftrightarrow \boldsymbol{m}_1\!=\!\boldsymbol{m}_2$.
	\end{enumerate}
\end{thm}

\begin{pf}
	\begin{enumerate}[label=\arabic*)]
		\setlength{\leftmargin}{1em}
		\setlength{\labelsep}{0em}
		\setlength{\itemindent}{1em}
		\setlength{\leftskip}{-2em}
		\item \textbf{Symmetry}\textup{:}
		\begin{equation}
		\begin{aligned}
		\mathrm{PBAGD}( \boldsymbol{m}_1,\boldsymbol{m}_2 )= &\sum\nolimits_{k = 1}^{2^n}\!\!\!\frac{PB_{\boldsymbol{m}_1}(\!A_k\!)\! +\! PB_{\boldsymbol{m}_2}(\!A_k\!)}{2}\!\log \! \frac{PB_{\boldsymbol{m}_1}(\!A_k\!) \!+ \!PB_{\boldsymbol{m}_2}(\!A_k\!)}{{2\!\sqrt {PB_{\boldsymbol{m}_1}(\!A_k\!)PB_{\boldsymbol{m}_2}(\!A_k\!)} }} \\
		= &\sum\nolimits_{k = 1}^{2^n} \!\!\! \frac{{PB_{\boldsymbol{m}_2}(\!A_k\!) \!+\! PB_{\boldsymbol{m}_1}(\!A_k\!)}}{2}\!\log \! {\frac{{PB_{\boldsymbol{m}_2}(\!A_k\!)\! +\! PB_{\boldsymbol{m}_1}(\!A_k\!)}}{{2\sqrt {PB_{\boldsymbol{m}_2}(\!A_k\!)PB_{\boldsymbol{m}_1}(\!A_k\!)} }}} \\
		= &\mathrm{PBAGD}\left( \boldsymbol{m}_2,\boldsymbol{m}_1 \right).
		\end{aligned}
		\label{eq28:PBAGD_Proof_3}
		\end{equation}
		Hence, the symmetry of the PBAGD is proven.
		\item \textbf{Nonnegativity}\textup{:}
		$\forall{A_k}\subseteq\Omega, PB_{\boldsymbol{m}_i}(A_k)\ge{0}$. According to the inequality of arithmetic-geometric mean, there is $PB_{\boldsymbol{m}_1}(A_k)+PB_{\boldsymbol{m}_2}(A_k)\ge{2}\sqrt{ PB_{\boldsymbol{m}_1}(A_k)PB_{\boldsymbol{m}_2}(A_k)}\ge{0}$. Then\textup{:}		
		\begin{equation}
		\hspace{-0.9cm}
		\begin{aligned}
		\mathrm{PBAGD}( \boldsymbol{m}_1,\boldsymbol{m}_2 ) =&\sum\nolimits_{k = 1}^{2^n} \!\!\! \frac{PB_{\boldsymbol{m}_1}(\!A_k\!) \!+\! PB_{\boldsymbol{m}_2}(\!A_k\!)}{2}
		\log \!{\frac{{PB_{\boldsymbol{m}_1}(\!A_k\!) \!+\! PB_{\boldsymbol{m}_2}(\!A_k\!)}}{{2\!\sqrt {PB_{\boldsymbol{m}_1}(\!A_k\!)PB_{\boldsymbol{m}_2}(\!A_k\!)} }}} \\
		\ge& \sum\nolimits_{k = 1}^{2^n} \!\!\!{\frac{PB_{\boldsymbol{m}_1}(A_k) \!+\! PB_{\boldsymbol{m}_2}(A_k)}{2}}\log(1)
		= 0.
		\end{aligned}
		\label{eq23:PBAGD_Proof_1}
		\end{equation}		
		Hence, the nonnegativity of the PBAGD is proven.
		\item \textbf{Nondegeneracy}\textup{:} On the one hand, 
		if $\boldsymbol{m}_1 = \boldsymbol{m}_2$, $\forall{A_k}\subseteq\Omega, PB_{\boldsymbol{m}_1}(A_k)=PB_{\boldsymbol{m}_2}(A_k)$ and $PB_{\boldsymbol{m}_1}(A_k) + PB_{\boldsymbol{m}_2}(A_k) = 2\sqrt{PB_{\boldsymbol{m}_1}(A_k) PB_{\boldsymbol{m}_2}(A_k)}$. Thus
		\begin{equation}
		\hspace{-0.9cm}
		\mathrm{PBAGD}( \boldsymbol{m}_1,\boldsymbol{m}_2 )
		= \sum\nolimits_{k = 1}^{{2^n}} \!\!\!\frac{PB_{\boldsymbol{m}_1}(\!A_k\!) + PB_{\boldsymbol{m}_2}(\!A_k\!)}{2} \log ( 1 )
		= 0.
		\label{eq24:PBAGD_Proof_2_1}
		\end{equation}		
		On the other hand, since $\forall{A_k}\subseteq\Omega, PB_{\boldsymbol{m}_i}(A_k)\ge{0}$, $\boldsymbol{m}_1=\boldsymbol{m}_2$ holds if $\mathrm{PBAGD}(\boldsymbol{m}_1, \boldsymbol{m}_2)=0$. Thus, $\forall{A_k}\!\in\!{2^{\Omega}}$, one of the following two equations is valid\textup{:}
		\begin{equation}
		{PB_{\boldsymbol{m}_1}(A_k) + PB_{\boldsymbol{m}_2}(A_k)}  = 0, \qquad \log{\frac{PB_{\boldsymbol{m}_1}\left(A_k\right)+PB_{\boldsymbol{m}_2}\left(A_k\right)}{2\sqrt{PB_{\boldsymbol{m}_1}\left(A_k\right)PB_{\boldsymbol{m}_2}\left(A_k\right)}}} = 0.
		\label{eq25:PBAGD_Proof_2_2}
		\end{equation}
		There is $PB_{\boldsymbol{m}_1}(A_k) = PB_{\boldsymbol{m}_2}(A_k) = 0$ or  $PB_{\boldsymbol{m}_1}(A_k) + PB_{\boldsymbol{m}_2}(A_k) = 2\sqrt{PB_{\boldsymbol{m}_1}(A_k)PB_{\boldsymbol{m}_2}(A_k)}$. Therefore, Eq.\textup{(\ref{eq25:PBAGD_Proof_2_2})} hold only $PB_{\boldsymbol{m}_1}( A_k ) = PB_{\boldsymbol{m}_2}( A_k )$.  According to Lemma~\ref{Lemma 1}, $\boldsymbol{m}_1 = \boldsymbol{m}_2$ holds.
		
		Hence, the nondegeneracy of the PBAGD is proven.
	\end{enumerate}
\end{pf}

Nondegeneracy ensures that the distance between an evidence and the event evidence is zero if and only if they are identical. Since the event evidence assigns its entire mass to the candidate event assumed to be the ground truth, a zero distance indicates that the evaluated evidence provides complete support for that event. Symmetry ensures that exchanging the positions of the evaluated evidence and the event evidence does not change their distance or the resulting event-conditioned credibility. Together with nonnegativity, these properties support the use of PBAGD in event-conditioned credibility calculation.

\begin{example}
	\label{ex6}
	Consider the following five evidences $\{\boldsymbol{m}_i\}_{i=1}^{5}$ defined on the FoD $\Omega\!\!=\!\!\{\tilde{A}_i\}_{i=1}^{4}$\textup{:}
	\begin{equation}
	\begin{aligned}
	&{m_1}(\!\{\!\tilde{A}_1\!\}\!)\!=\!0.75,
	{m_1}(\!\{\!\tilde{A}_2\!\}\!)\!=\!
	{m_1}(\!\{\!\tilde{A}_3\!\}\!)\!=\!0.10,
	{m_1}(\!\Omega\!)\!             =\!0.05;\\
	&{m_2}(\!\{\!\tilde{A}_1\!\}\!)\!=\!0.65,
	{m_2}(\!\{\!\tilde{A}_2\!\}\!)\!=\!
	{m_2}(\!\{\!\tilde{A}_3\!\}\!)\!=\!0.10,
	{m_2}(\!\Omega\!)\!             =\!0.15;\\
	&{m_3}(\!\{\!\tilde{A}_1\!\}\!)\!=\!0.75,
	{m_3}(\!\{\!\tilde{A}_2\!\}\!)\!=\!
	{m_3}(\!\{\!\tilde{A}_3\!\}\!)\!=\!0.10,
	{m_3}(\!\Omega\!)\!             =\!0.05;\\
	&{m_4}(\!\{\!\tilde{A}_1\!\}\!)\!=\!0.75,
	{m_4}(\!\{\!\tilde{A}_2\!\}\!)\!=\!
	{m_4}(\!\{\!\tilde{A}_3\!\}\!)\!=\!0.10,
	{m_4}(\!\{\!\tilde{A}_4\!\}\!)\!=\!0.05;\\
	&{m_5}(\!\{\!\tilde{A}_1\!\}\!)\!=\!0.75,
	{m_5}(\!\{\!\tilde{A}_2\!\}\!)\!=\!
	{m_5}(\!\{\!\tilde{A}_3\!\}\!)\!=\!0.10,
	{m_5}(\!\{\!\tilde{A}_4\!\}\!)\!=\!0.05.
	\end{aligned}
	\notag
	\end{equation}
\end{example}
Applying PBAGD to Example \ref{ex6} yields: $\mathrm{PBAGD}(\!\boldsymbol{m}_1,\!\boldsymbol{m}_2\!)\!=\!\mathrm{PBAGD}(\!\boldsymbol{m}_2,\!\boldsymbol{m}_1\!)\!=\!0.0088$, confirming the symmetry property. Furthermore, $\mathrm{PBAGD}(\!\boldsymbol{m}_1,\!\boldsymbol{m}_3\!)\!=\!\mathrm{PBAGD}(\!\boldsymbol{m}_4,\!\boldsymbol{m}_5\!)\!=\!0$, which satisfies the nondegeneracy as $\boldsymbol{m}_1\!\!=\!\!\boldsymbol{m}_3$ and $\boldsymbol{m}_4\!=\!\boldsymbol{m}_5$. This holds true regardless of whether the focal elements are singletons or compound sets, demonstrating the generality of the property. The nonnegativity of PBAGD is also evident from all results being greater than or equal to zero.

\begin{figure*}[!h]
	\centering
	\begin{minipage}{0.49\linewidth}
		\centering
		\includegraphics[width=3.2in]{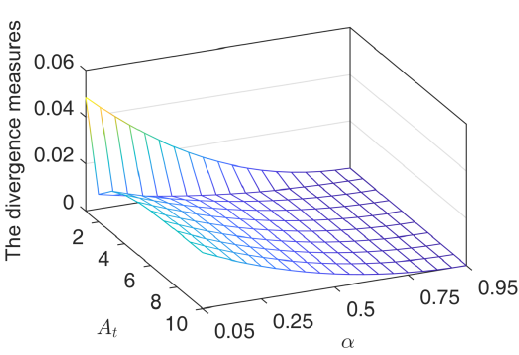}
		\caption{Results for the PBAGD.}
		\label{fig3a: PBAGD_Curve_3D}
	\end{minipage}
	\begin{minipage}{0.49\linewidth}
		\centering
		\includegraphics[width=3.2in]{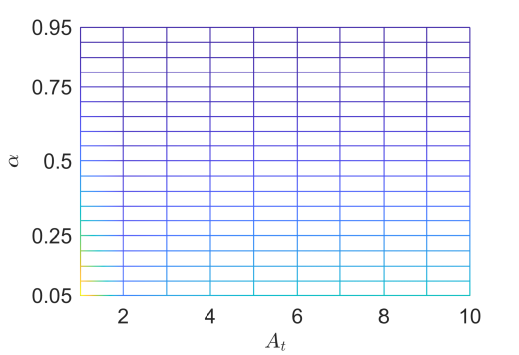}
		\caption{Variation of $\alpha$ with varying $A_t$.}
		\label{fig3b: PBAGD_Curve_At_alpha}
	\end{minipage}
	\begin{minipage}{0.49\linewidth}
		\centering
		\includegraphics[width=3.2in]{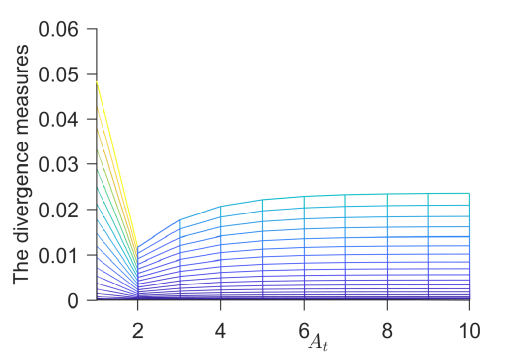}
		\caption{Variation of PBAGD with varying $A_t$.}
		\label{fig3c: PBAGD_Curve_At_Div}
	\end{minipage}
	\begin{minipage}{0.49\linewidth}
		\centering
		\includegraphics[width=3.2in]{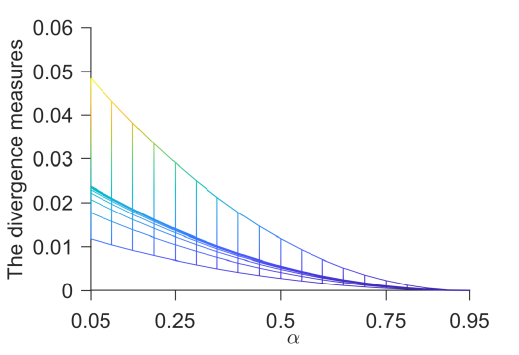}
		\caption{Variation of PBAGD with varying $\alpha$.}
		\label{fig3d: PBAGD_Curve_alpha_Div}
	\end{minipage}
\end{figure*}

\begin{example}
	\label{ex2}
	Consider the following two evidences $\{\boldsymbol{m}_i\}_{i=1}^{2}$ defined on $\Omega\!\!=\!\!\{\tilde{A}_i\}_{i=1}^{11}$\textup{:}
	\begin{equation}
	{m_1}({\{\tilde{A}_2\}}) = \alpha,{m_1}(A_t) = 1 - \alpha;\quad
	{m_2}({\{\tilde{A}_2\}}) = 0.95,{m_2}(A_t) = 0.05.
	\nonumber
	\end{equation}
	where $\alpha \in [0.05,0.95]$, $A_t=\{{\tilde{A}_i}\}_{i=1}^{t},t=1,2,\cdots,10$.
\end{example}
Figs.\ref{fig3a: PBAGD_Curve_3D}-\ref{fig3d: PBAGD_Curve_alpha_Div} show the variation of PBAGD with $A_t$ and $\alpha$ for Example \ref{ex2}. Figs.\ref{fig3a: PBAGD_Curve_3D}-\ref{fig3b: PBAGD_Curve_At_alpha} confirm PBAGD is nonnegative across all parameter variations. In Fig.\ref{fig3c: PBAGD_Curve_At_Div}, PBAGD peaks at $t=1$ for $\alpha=0.05$, and increases with $t$ from 2 to 10, consistent with the declining similarity between $A_t$ and $\{{\tilde{A}_2}\}$ as more events are included in $A_t$. Fig.\ref{fig3d: PBAGD_Curve_alpha_Div} shows PBAGD decreasing to 0 as $\alpha$ approaches 0.95, where $\boldsymbol{m}_1 = \boldsymbol{m}_2$.

\begin{example}
	\label{ex3}
	Consider the following two evidences $\{\boldsymbol{m}_i\}_{i=1}^{2}$ defined on $\Omega=\{\tilde{A}_i\}_{i=1}^{11}$, with $A_t=\{{\tilde{A}_i}\}_{i=1}^{t},t=1,2,\cdots,10$\textup{:}
	\begin{equation}
	{m_1}( \Omega )={m_1}(\{\tilde{A}_7\})= 0.10,{m_1}(\{\tilde{A}_i\}_{i=2}^{4})=0.05,{m_1}(A_t) = 0.80; \qquad
	{m_2}(\{\tilde{A}_i\}_{i=1}^{5}) = 1.
	\notag
	\end{equation}
\end{example}

\begin{figure*}[!h]
	\centering
	\begin{minipage}{0.45\linewidth}
		\centering
		\includegraphics[width=3.1in]{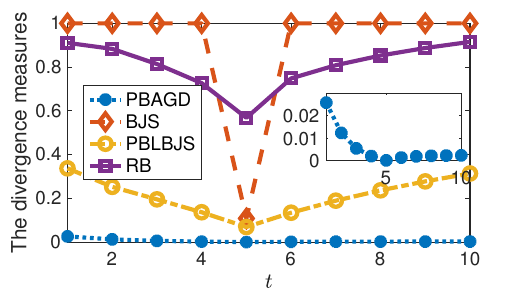}
		\caption{Variation of 4 EDMFs with $A_t$.}
		\label{fig4: DivCompare_ABCDE}
	\end{minipage}
	\begin{minipage}{0.53\linewidth}
		\centering
		\includegraphics[width=3.1in]{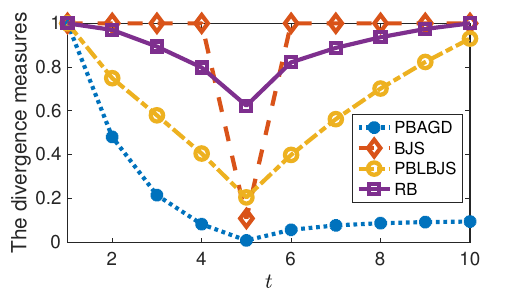}
		\caption{Variation of maximized 4 EDMFs with $A_t$.}
		\label{fig5: DivCompare_ABCDE}
	\end{minipage}
\end{figure*}
Fig.\ref{fig4: DivCompare_ABCDE} compares four EDMFs for evidences $\boldsymbol{m}_1$ and $\boldsymbol{m}_2$ in Example \ref{ex3} as $t$ varies from 1 to 10. BJS remains nearly constant except at $t=5$, whereas the other measures decrease to a minimum at $t=5$, where both evidences assign their largest mass to $\{\tilde{A}_i\}_{i=1}^5$, and then increase. The curves of BJS, RB, and PBLBJS are approximately symmetric around $t=5$. PBAGD exhibits an asymmetric response to the increasing cardinality of the compound focal element in this example. This response follows from its joint use of belief and plausibility values over the complete power set, through which changes in support conveyed by singleton and compound focal elements affect the divergence differently. Fig.\ref{fig5: DivCompare_ABCDE} further shows that PBAGD changes more sharply than the other EDMFs as $t$ increases from 1 to 5. Through the nonincreasing kernel in Eq.\eqref{eq:ecc_general_kernel}, this difference affects the event-conditioned credibility assigned to the evaluated evidence. These observations characterize the behavior of PBAGD in the present example; its effect within the joint fusion model is evaluated in Section~\ref{sec:Experiment and application}.

\section{Decision-Oriented Joint Optimization for Evidence Fusion}
\label{sec:Iterative credible evidence fusion}
Based on event-conditioned credibility, this section formulates JOFM and presents a relaxed fixed-point iteration as its default solver, with Kuhn simplicial search used if the iteration enters a periodic orbit.

\subsection{Joint Optimization Model}
\label{subsec:joint_optimization_model}
As reviewed in Section~\ref{sec: Problem formulation}, existing CEF methods
generally share three basic components: credibility-based evidence
preprocessing, evidence combination using DCR, and decision making based on
the pignistic probabilities derived from the fusion result. JOFM retains these
components and couples credibility calculation, evidence fusion, and event
decision through event-conditioned credibility.

For fusion result $\boldsymbol{m}_{\oplus}$, the Pignistic probability of candidate event $\tilde{A}_j$ is calculated as\cite{21_liu2011combination}
\begin{equation}
p(\tilde{A}_j)=BetP_{\boldsymbol{m}_{\oplus}}(\tilde{A}_j)=\sum\nolimits_{{\tilde{A}_j}\subseteq{B}\subseteq\Omega}\frac{m_{\oplus}(B)}{|B|}
\label{eq:candidate_event_probability}
\end{equation}
where $p(\tilde{A}_j)$ is the probability that $\tilde{A}_j$ is ground truth. Under the MPPDR, the decision is given by $\tilde{A}^{*}=\arg\max_{\tilde{A}_j\in\Omega} p(\tilde{A}_j)$. Eq.\eqref{eq:candidate_event_probability} maps the $\boldsymbol{m}_{\oplus}$ back to the candidate-event probabilities, yielding the following joint model:

\begin{align}
\label{eq:JOFM_summary_Eq_1} Cred_i&=\sum\nolimits_{j=1}^n p(\phi_i|\tilde{A}_j)p(\tilde{A}_j)\\
\label{eq:JOFM_summary_Eq_3} \boldsymbol{m}_{avg}&=\sum\nolimits_{i=1}^N {Cred_i \boldsymbol{m}_i}
\\
\label{eq:JOFM_summary_Eq_2} \boldsymbol{m}_{\oplus}&=\bigoplus\nolimits_{l=1}^{N}\boldsymbol{m}_{avg}\\
\label{eq:JOFM_summary_Eq_4} p(\tilde{A}_j)&=BetP_{{\boldsymbol{m}_{\oplus}}}(\tilde{A}_j)
\end{align}

Let $\boldsymbol{p}=[p(\tilde{A}_1),\ldots,p(\tilde{A}_n)]^{\mathrm T}$, and
$\boldsymbol{\mathrm{BetP}}(\boldsymbol{m})
=[\mathrm{BetP}_{\boldsymbol{m}}(\tilde{A}_1),\ldots,
\mathrm{BetP}_{\boldsymbol{m}}(\tilde{A}_n)]^{\mathrm{T}}$. Substituting Eqs.\eqref{eq:JOFM_summary_Eq_1}--\eqref{eq:JOFM_summary_Eq_4} into a single mapping gives
\begin{equation}
\boldsymbol{p}=\mathcal{F}(\boldsymbol{p})=\boldsymbol{\mathrm{BetP}}\!\left(\bigoplus\nolimits_{l=1}^{N}(\sum\nolimits_{i=1}^{N}\sum\nolimits_{j=1}^{n}
p(\phi_i\mid\tilde{A}_j)p(\tilde{A}_j)\boldsymbol{m}_i)\right).
\label{eq:fixed_point_model}
\end{equation}
Here, $\mathcal{F}$ compactly represents the coupled computations, rather than
an additional fusion operation. Solving JOFM therefore is finding
$\boldsymbol{p}$ satisfying the fixed-point equation
$\boldsymbol{p}=\mathcal{F}(\boldsymbol{p})$.

The vector $\boldsymbol{p}$ belongs to the probability simplex 	$\Delta^{n-1}=\{\boldsymbol{p}\in\mathbb{R}^{n}:p_j\geq0,\sum_{j=1}^{n}p_j=1\}$. For fixed event-conditioned credibilities, their columnwise normalization implies that the overall credibilities given by Eq.\eqref{eq:JOFM_summary_Eq_1} are nonnegative and sum to one. Hence, $\boldsymbol{m}_{\mathrm{avg}}$ is a normalized mass function. Moreover, for some nonempty focal element $B$, $m_{\mathrm{avg}}(B)>0$, so its $N$-fold conjunctive combination satisfies $1-K\geq m_{\mathrm{avg}}(B)^N>0$. DCR is therefore well defined. Since credibility aggregation and average-evidence construction are linear in $\boldsymbol{p}$, while conjunctive combination, DCR normalization, and the Pignistic transformation are continuous, $\mathcal{F}$ is a continuous self-map of $\Delta^{n-1}$. By the Brouwer fixed-point theorem, Eq.\eqref{eq:fixed_point_model} has at least one solution. It makes fixed-point computation equivalent to the residual minimization problem:

\begin{equation}
\min_{\boldsymbol{p}\in\Delta^{n-1}}r(\boldsymbol{p}),
\qquad
r(\boldsymbol{p})
=\left\|\mathcal{F}(\boldsymbol{p})-\boldsymbol{p}\right\|_1.
\label{eq22:Termination_aim_opti}
\end{equation}
Since Eq.\eqref{eq:fixed_point_model} has at least one solution, the global minimum of Eq.\eqref{eq22:Termination_aim_opti} is zero. The joint optimization in this paper refers to finding candidate-event probabilities that reduce the fixed-point residual to zero under the coupled relationship defined by Eqs.\eqref{eq:JOFM_summary_Eq_1}--\eqref{eq:JOFM_summary_Eq_4}.

\begin{remark}
	In labeled examples, critical and anomalous evidence are defined ex post relative to the known ground-truth event. In an unlabeled decision task, JOFM does not prespecify either category. Instead, it evaluates each candidate event as a hypothesis and jointly determines candidate-event probabilities and the associated credibility allocation. Therefore, the operational output represents decision-consistent credibility, whereas the critical/anomalous terminology is used only for post hoc interpretation if the true event is available.
\end{remark}
\subsection{Numerical Solution of the Joint Model}
\label{subsec:numerical_solution}

\subsubsection{Fast Solution by Relaxed Fixed-Point Iteration}

As the ground truth is unknown and no additional prior information is available, all candidate events have equal probabilities at the beginning of the solution process
\begin{equation}
p^{(0)}(\tilde{A}_j)=\frac{1}{n},\quad j=1,\ldots,n.
\label{eq20:PatternPriorProb_Avg}
\end{equation}

Given the event probabilities $\boldsymbol{p}^{(k)}$ at iteration $k$, the mapping value $\mathcal{F}(\boldsymbol{p}^{(k)})$ is calculated according to the JOFM. The relaxed fixed-point update is
\begin{equation}
\boldsymbol{p}^{(k+1)}
=(1-\epsilon)\boldsymbol{p}^{(k)}
+\epsilon\mathcal{F}(\boldsymbol{p}^{(k)}),
\qquad 0<\epsilon\leq1.
\label{eq:relaxed_fixed_point_iteration}
\end{equation}
For any relaxation factor $\epsilon>0$, Eqs.\eqref{eq:relaxed_fixed_point_iteration} and \eqref{eq:fixed_point_model} have the same fixed points. The relaxation factor therefore changes only the numerical trajectory and not the fixed-point set of the JOFM.

The fast solver uses $\epsilon=1$ by default, in which case Eq.\eqref{eq:relaxed_fixed_point_iteration} reduces to direct fixed-point iteration without introducing an additional parameter. The iteration terminates if $r(\boldsymbol{p}^{(k)})\leq\delta$, indicating that the fixed-point relationship is satisfied within the prescribed residual tolerance. Moreover, Eq.\eqref{eq:relaxed_fixed_point_iteration} gives
\begin{equation}
\|\boldsymbol{p}^{(k+1)}-\boldsymbol{p}^{(k)}\|_1=\epsilon{r}(\boldsymbol{p}^{(k)}).
\label{eq:relaxed_step_residual}
\end{equation}
For $\epsilon=1$, the fixed-point residual equals the difference between two consecutive probability vectors. For $0<\epsilon<1$, however, this difference is only $\epsilon$ times the residual. Therefore, $r(\boldsymbol{p}^{(k)})\leq\delta$ is retained as the termination criterion for the relaxed iteration to avoid premature termination caused by a small $\epsilon$.

Since Eq.\eqref{eq:relaxed_fixed_point_iteration} always maps the event probabilities into the probability simplex, the iterative states remain bounded and cannot diverge without bound in norm. If direct iteration does not reach the residual threshold, it may be converging slowly or may have entered a bounded periodic orbit. If the residual continues to decrease steadily when the prescribed iteration limit is reached, the computational budget is increased. If the residual remains above the threshold and the event probabilities repeatedly return to nonconsecutive historical states, the iteration may have entered a periodic orbit. Specifically, if there exists an integer $q\geq2$ such that $\|\boldsymbol{p}^{(k)}-\boldsymbol{p}^{(k-q)}\|_1\leq\delta$, and this relation recurs in subsequent iterations, $q$ is treated as a detected candidate period. The same $L_1$ norm and threshold are used for periodic-orbit detection and the fixed-point residual, so no additional detection parameter is introduced.

For a detected periodic orbit, choosing $0<\epsilon<1$ reduces the magnitude of each feedback update and adds numerical damping. Relaxation alone, however, provides no general convergence guarantee. This paper therefore uses $\epsilon=1$ as the default setting for fast computation and makes no general claim regarding the convergence or optimality of other values of $\epsilon$. If direct iteration enters a periodic orbit, the Kuhn simplicial method presented next is used to search for an approximate fixed point.

\begin{algorithm}[!t]
	\caption{Fast Fixed-Point Solver for joint model.}
	\label{alg1:FeecbackFusionAlgorithm}
	\begin{algorithmic}
		\REQUIRE $\{\boldsymbol{m}_i\}_{i=1}^{N}$, $\Omega=\{\tilde{A}_j\}_{j=1}^{n}$, termination threshold $\delta$, maximum iteration number $Iter_{\max}$.
		\ENSURE Fusion result $\boldsymbol{m}_{\oplus}$.
		\STATE Construct the event evidences and calculate the event-conditioned credibility matrix.
		\STATE Initialize $\boldsymbol{p}^{(0)}$ according to Eq.\eqref{eq20:PatternPriorProb_Avg}.
		\FOR{$k=0,1,\ldots,Iter_{\max}-1$}
		\STATE Calculate $\mathcal{F}(\boldsymbol{p}^{(k)})$ and the corresponding fusion result according to Eqs.\eqref{eq:JOFM_summary_Eq_1}--\eqref{eq:JOFM_summary_Eq_4}.
		\IF{$r(\boldsymbol{p}^{(k)})\leq\delta$}
		\STATE \textbf{break}
		\ENDIF
		\STATE $\boldsymbol{p}^{(k+1)}\leftarrow\mathcal{F}(\boldsymbol{p}^{(k)})$.
		\ENDFOR
		\STATE Return the fusion result $\boldsymbol{m}_{\oplus}$ from the last evaluation of the joint model.
	\end{algorithmic}
\end{algorithm}

\subsubsection{Fixed-Point Search by the Kuhn Simplicial Method}

As established above, $\mathcal{F}$ continuously maps the probability simplex into itself, and the JOFM has at least one fixed point. Thus, the Kuhn simplicial method uses triangulation and combinatorial labeling to search for an approximate fixed point without requiring convergence of the direct iterative trajectory~\cite{kuhn1968simplicial}. In the present model, the search variable is the candidate-event probability vector $\boldsymbol{p}$ rather than the mass function $\boldsymbol{m}$. Compound focal elements participate in the internal evaluation of $\mathcal{F}(\boldsymbol{p})$ but do not change its probability-simplex domain.

For a prescribed maximum mesh diameter $h$, the probability simplex $\Delta^{n-1}$ is triangulated, and each mesh vertex $\boldsymbol{v}$ is assigned a standard Sperner label according to $\boldsymbol{v}-\mathcal{F}(\boldsymbol{v})$. The implementation used in this paper evaluates every mesh vertex, enumerates the simplices in the finite triangulation, and identifies all completely labeled simplices. For each such simplex, a piecewise-linear candidate is calculated from its vertex coordinates and mapping differences. Candidates outside the simplex because of finite-mesh interpolation error are replaced by the lowest-residual vertex of that simplex. Duplicate candidates are clustered, and the candidate with the smallest fixed-point residual is retained as $\hat{\boldsymbol{p}}_{h}$. This exhaustive finite-mesh implementation differs from following a single simplicial path, but it uses the same Kuhn triangulation and Sperner-labeling construction.

The accuracy of a finite-mesh candidate is restricted by the mesh size. If $r(\hat{\boldsymbol{p}}_{h}) \leq \delta$, the current candidate is accepted. Otherwise, $\hat{\boldsymbol{p}}_{h}$ is used as the initial point for numerical refinement of the fixed-point equation within the neighborhood identified by the candidate simplex. Since the event probabilities satisfy the normalization constraint, only $n-1$ independent variables need to be considered. The reduced fixed-point equation is written as
\begin{equation}
\mathcal{G}(\boldsymbol{x})= [\mathcal{F}_{1}(\boldsymbol{p})-p_{1}, \cdots, \mathcal{F}_{n-1}(\boldsymbol{p})-p_{n-1}]^T =\boldsymbol{0},\quad {p}_n=1-\sum\nolimits_{j=1}^{n-1}p_j,
\label{eq:kuhn_local_refinement}
\end{equation}
where $\boldsymbol{x}=[p_1,\ldots,p_{n-1}]^{\mathrm T}$. In the numerical implementation, Eq.\eqref{eq:kuhn_local_refinement} is solved by MATLAB's \texttt{fsolve} with the Levenberg--Marquardt algorithm, initialized at $\hat{\boldsymbol{p}}_{h}$. The result is accepted only if all event probabilities are nonnegative, sum to one within numerical tolerance, and satisfy the prescribed fixed-point residual threshold. If these conditions are not met within the computational budget, the mesh is refined and the simplicial search is repeated. This local numerical step reduces the discretization error of the finite mesh; it does not replace the Kuhn search or change the candidate region identified by the completely labeled simplex.

Since a continuous mapping on a compact simplex is uniformly continuous, the diameter of a completely labeled simplex and the corresponding fixed-point approximation error approach zero as $h\rightarrow{0}$. The Kuhn method can therefore approximate a fixed point to any prescribed residual accuracy through mesh refinement. Refinement within the candidate region reduces the global mesh resolution required to reach the prescribed accuracy; it does not imply that an exact fixed point is obtained on an arbitrary finite mesh.

The finite-mesh search may produce multiple distinct candidates satisfying the simplex constraints and the prescribed residual threshold. If their MPPDR decisions are identical, the common decision is reported, and the candidate with the smallest residual is retained as the numerical representative. If the candidates support different decision events, decision ambiguity is reported and the corresponding candidates are retained without forcing a single decision. This rule applies only to the candidates found under the prescribed finite search budget and does not imply enumeration of all mathematical fixed points.

This paper assumes that only one basic event actually occurs. If multiple events can occur simultaneously or the problem is formulated in an open-world setting, the candidate decision events, event-conditioned credibility, and decision transformation must be redefined. Such extensions are beyond the scope of this paper.

\subsection{Complexity and Scalability Analysis}
\label{Complexity and scalability analysis}

PBAGD computes the event-conditioned credibility matrix once before fusion by evaluating the EDMs between $N$ evidences and $n$ event evidences over the $2^n$ subsets of the FoD, resulting in a complexity of $O(nN2^n)$. Each subsequent mapping evaluation requires $O(nN)$ operations to calculate the overall credibility and $O(N2^n)$ operations to construct the $\boldsymbol{m}_{\mathrm{avg}}$. Sequentially performing $N-1$ DCR would additionally require $O((N-1)4^n)$ operations.

Since the JOFM combines $N$ identical copies of the $\boldsymbol{m}_{\mathrm{avg}}$, repeated DCR combinations can be avoided using the multiplicative property of the commonality function. Let $Q_{\mathrm{avg}}$ be the commonality function of $\boldsymbol{m}_{\mathrm{avg}}$. The unnormalized conjunctive result satisfies
\begin{equation}
Q_{\cap}(A)=Q_{\mathrm{avg}}(A)^N,
\qquad A\subseteq\Omega.
\label{eq:commonality_power}
\end{equation}
The unnormalized mass function $\boldsymbol{m}_{\cap}$ is obtained by the inverse M\"obius transform, and
\begin{equation}
m_{\oplus}(A)=\frac{m_{\cap}(A)}{1-m_{\cap}(\varnothing)},
\quad A\neq\varnothing.
\label{eq:commonality_normalization}
\end{equation}

Precomputing the M\"obius transformation matrices avoids repeated matrix construction during conversion between commonality and mass function. Each dense transformation matrix has dimension $2^n\times2^n$, resulting in $O(4^n)$ operations, whose nonzero pattern is determined by subset inclusion. Since the number of ordered subset-inclusion pairs is $\sum_{B\subseteq\Omega}2^{|B|}=3^n$, sparse matrix--vector multiplication requires $O(3^n)$ operations. Fast M\"obius transforms on the Boolean subset lattice further reduce each forward or inverse transform to $O(n2^n)$ operations~\cite{kennes1992computational}, while the elementwise exponentiation in Eq.\eqref{eq:commonality_power} requires $O(2^n)$ operations. Consequently, one fast mapping evaluation has complexity $O(nN+N2^n+n2^n)=O((N+n)2^n)$. For $K_{\mathrm{iter}}$ iterations, the total complexity is $O\!\left(nN2^n+K_{\mathrm{iter}}(N+n)2^n\right)$, where the $O(n)$ costs of the relaxed update and residual calculation do not affect this bound.

For the Kuhn method, let $M_h$ denote the number of mapping evaluations at a prescribed mesh resolution. The simplicial search requires approximately $O(M_h(N+n)2^n)$ operations. If local numerical refinement requires another $K_{\mathrm{loc}}$ mapping evaluations, the total complexity becomes $O\!\left((M_h+K_{\mathrm{loc}})(N+n)2^n\right)$.

Relaxed fixed-point iteration is the default fast solver. The Kuhn method provides a fixed-point approximation with a residual guarantee under mesh refinement. If the number of evidences $N$ is large, exponentiation of the commonality function eliminates the multiplicative cost in $N$ caused by repeated DCR, although constructing the average evidence retains linear complexity in $N$.

\section{Experiments and Analysis}
\label{sec:Experiment and application}

This section evaluates the proposed method in terms of credibility assessment, numerical solution of the joint model, computational efficiency, and applicability to different evidence dissimilarity measures.

\subsection{Experimental Overview and Common Settings}

The experiments are organized into four subsections. Section~\ref{subsec: Credibility Assessment of Critical and Anomalous Evidences} evaluates whether JOFM effectively utilizes critical evidence while suppressing anomalous evidence. Section~\ref{subsec: Numerical Solution of the Joint Model} investigates the numerical solution of the joint model, including the empirical fixed-point attainment of the fast solver, relaxation of periodic iteration, and supplementary fixed-point search using the Kuhn simplicial method. Section~\ref{subsec: Computational Efficiency and Scalability} evaluates computational efficiency through runtime comparisons under a common mapping-evaluation budget and acceleration using the commonality function. Section~\ref{subsec: Applicability to Different Evidence Dissimilarity Measures} examines the applicability of JOFM to different EDMFs.

Unless otherwise specified, PBAGD is used as the EDMF. In terms of the kernel function, the experiments in Sections \ref{subsec: Credibility Assessment of Critical and Anomalous Evidences}, \ref{subsec: Computational Efficiency and Scalability} and \ref{subsec: Applicability to Different Evidence Dissimilarity Measures} use Eq.\eqref{eq:inverse_kernel}, while Section \ref{subsec: Numerical Solution of the Joint Model} uses Eq.\eqref{eq:exponential_kernel}. Section \ref{subsubsec: Empirical Attainment Performance of Fast Fixed-Point Iteration} considers different $\tau$ values, while Sections \ref{subsubsec: Periodic Orbit and Relaxed Updates} and \ref{subsubsec: Kuhn Simplicial Fixed-Point Search} use $\tau = 1000$. Except for the relaxation-factor analysis in Section~\ref{subsubsec: Periodic Orbit and Relaxed Updates}, the fast fixed-point solver uses $\epsilon=1$, a residual threshold of $\delta=10^{-8}$, and a maximum of 300 iterations.

All experiments are conducted using MATLAB R2020a on a computer equipped with an AMD 4800H processor and 32~GB of RAM. In the runtime experiments, each of the 30 independently generated inputs is executed five times, and the median runtime is recorded.

\subsection{Credibility Assessment of Critical and Anomalous Evidences}
\label{subsec: Credibility Assessment of Critical and Anomalous Evidences}

The two examples in Section~\ref{sec: Problem formulation} are used to evaluate the JOFM in terms of credibility assessment and fusion performance. The credibility update process is first examined using Example~\ref{ex1}, after which the fusion results of the JOFM and representative CEF methods are compared for both examples. Because the ground-truth events are specified in both examples, critical and anomalous evidence are used here as ex post evaluation labels.

\begin{table}[!h]
	\setlength\tabcolsep{1.5mm}
	\caption{The credibility of evidence in Example \ref{ex1}.\label{tab4:Record_Cred_Compare_Prior}}
	\centering
	\renewcommand{\arraystretch}{1}
	\begin{tabular}{lcccccccccc}
		\toprule[1.25pt]
		\multirow{2}{*}{Evidence} & \multicolumn{10}{c}{Iteration} \\
		\cline{2-11}
		& 1 & 2 & 3 & 4 & 5 & 6 & 7 & 8 & 9 & 10 \\
		\midrule[0.5pt]
		$\boldsymbol{m}_1$ & 0.153 & 0.184 & 0.210 & 0.211 & 0.211 & 0.211 & 0.211 & 0.211 & 0.211 & 0.211 \\
		$\boldsymbol{m}_2$ & 0.170 & 0.248 & 0.285 & 0.288 & 0.288 & 0.288 & 0.288 & 0.288 & 0.288 & 0.288 \\
		$\boldsymbol{m}_3$ & 0.139 & 0.125 & 0.138 & 0.139 & 0.139 & 0.139 & 0.139 & 0.139 & 0.139 & 0.139 \\
		$\boldsymbol{m}_4$ & 0.188 & 0.300 & 0.348 & \textbf{0.351} & \textbf{0.351} & \textbf{0.351} & \textbf{0.351} & \textbf{0.351} & \textbf{0.351} & \textbf{0.351} \\
		$\boldsymbol{m}_5$ & 0.349 & 0.144 & 0.019 & 0.011 & 0.011 & 0.011 & 0.011 & 0.011 & 0.011 & 0.011 \\
		\bottomrule[1.25pt]
	\end{tabular}
\end{table}
Under uniform initialization, Table~\ref{tab4:Record_Cred_Compare_Prior} shows that the credibility ordering reverses and stabilizes within four iterations, i.e., critical evidence $\boldsymbol{m}_4$ becomes the most credible evidence, whereas anomalous evidence $\boldsymbol{m}_5$ becomes the least credible. This reversal results from incorporating the fusion-derived event probabilities into subsequent credibility calculations.

\begin{table}[!h]
	\setlength\tabcolsep{0.7mm
		\caption{Fusion results of 5 evidences in Table \ref{tab1:Example1Evidence}.\label{tab5:Fus_Res_tab1}}
		\centering
		\renewcommand{\arraystretch}{1}
		\begin{tabular}{ccccc|ccccc}
			\toprule[1.25pt]
			Method& $\{\!\tilde{A}_1\!\}$ & $\{\!\tilde{A}_2\!\}$ & $\{\!\tilde{A}_3\!\}$ & $\Omega$&Method& $\{\!\tilde{A}_1\!\}$ & $\{\!\tilde{A}_2\!\}$ & $\{\!\tilde{A}_3\!\}$ & $\Omega$ \\
			\midrule[0.5pt]
			DCR\cite{11_dempster2008upper}     &0.0000&0.3443&0.6557&0.0000&
			Murphy\cite{19_murphy2000combining}&0.9715&0.0055&0.0222&0.0008\\
			Dismp\cite{21_liu2011combination}  &0.9833&0.0082&0.0032&0.0053&
			BJS\cite{16_xiao2019multi}         &0.9937&0.0030&0.0025&0.0008\\
			RB\cite{22_xiao2020new}            &0.9914&0.0034&0.0043&0.0008&
			PBLBJS\cite{28_wang2021new}        &0.9957&0.0026&0.0008&0.0009\\
			RFBD\cite{zhang2025reinforced}     &0.9892&0.0065&0.0008&0.0035&
			PBAGD        &0.9942&0.0029  &0.0021&0.0008\\
			JOFM-PBAGD   &\textbf{0.9973}&0.0012&0.0006&0.0009&&&&&\\
			\bottomrule[1.25pt]
	\end{tabular}}
\end{table}

Table~\ref{tab5:Fus_Res_tab1} shows that DCR incorrectly favors $\tilde{A}_3$ under the high conflict caused by $\boldsymbol{m}_5$, whereas all the compared CEF methods identify $\tilde{A}_1$ under MPPDR. JOFM-PBAGD provides the strongest support for $\tilde{A}_1$, with a mass of 0.9973.

\begin{table}[!h]
	\setlength\tabcolsep{0.7mm
		\caption{Fusion results of 5 evidences in Table \ref{tab5:Example1Credibility_modify}.\label{tab6:Fus_Res_tab4}}
		\centering
		\renewcommand{\arraystretch}{1}
		\begin{tabular}{ccccc|ccccc}
			\toprule[1.25pt]
			Method& $\{\!\tilde{A}_1\!\}$ & $\{\!\tilde{A}_2\!\}$ & $\{\!\tilde{A}_3\!\}$ & $\Omega$&Method& $\{\!\tilde{A}_1\!\}$ & $\{\!\tilde{A}_2\!\}$ & $\{\!\tilde{A}_3\!\}$ & $\Omega$ \\
			\midrule[0.5pt]
			DCR\cite{11_dempster2008upper}     &0.0000&0.5180&0.4820&0.0000&
			Murphy\cite{19_murphy2000combining}&0.6859&0.1453&0.1641&0.0047\\
			Dismp\cite{21_liu2011combination}  &0.6608&0.1835&0.1225&0.0332&
			BJS\cite{16_xiao2019multi}         &0.7852&0.1240&0.0847&0.0061\\
			RB\cite{22_xiao2020new}            &0.7664&0.1290&0.0989&0.0057&
			PBLBJS\cite{28_wang2021new}        &0.7805&0.1423&0.0679&0.0093\\
			RFBD\cite{zhang2025reinforced}     &0.7833&0.1346&0.0662&0.0159&
			PBAGD        &0.7619&0.1390  &0.0921&0.0071\\
			JOFM-PBAGD   &\textbf{0.9702}&0.0216&0.0063&0.0019&&&&&\\
			\bottomrule[1.25pt]
	\end{tabular}}
\end{table}
For the more uncertain evidence set in Table~\ref{tab4:Example1Evidence_modify}, JOFM-PBAGD assigns a mass of 0.9702 to $\tilde{A}_1$, compared with 0.6608--0.7852 for the representative CEF methods in Table~\ref{tab6:Fus_Res_tab4}. It indicates that the proposed method more fully uses the information provided by the critical evidence $\boldsymbol{m}_4$ while reducing the influence of the anomalous evidence $\boldsymbol{m}_5$.

\subsection{Numerical Solution of the Joint Model}
\label{subsec: Numerical Solution of the Joint Model}
\subsubsection{Empirical Attainment Performance of Fast Fixed-Point Iteration}
\label{subsubsec: Empirical Attainment Performance of Fast Fixed-Point Iteration}

This Monte Carlo experiment characterizes the numerical behavior of the fast fixed-point solver under varying $n$, $N$, $\tau$, and evidence sparsity. Because the randomly generated evidence sets are not assigned ground-truth events, numerical performance is evaluated by whether the residual threshold in Eq.\eqref{eq22:Termination_aim_opti} is attained, rather than by decision correctness. It considers $n\in\{3,4,5\}$ basic events, $N\in\{3,5,10,20\}$ evidences, and $\tau\in\{0.1,1,10,100,1000\}$. For each parameter combination, 1000 dense and 1000 sparse evidence sets are generated, yielding 120,000 trials. For each dense evidence, mass function assigns randomly generated positive masses to $2^\Omega/\{\emptyset\}$, followed by normalization. In a sparse evidence, mass function contains 2--6 randomly selected nonempty focal elements, over which randomly generated positive masses are normalized.

The reported metrics are the proportion of trials attaining the residual threshold within 300 iterations, the proportion attaining it within 10 iterations, the median iteration count among attained trials, and the proportion remaining unattained after iteration $k$.

\begin{sidewaysfigure}[htbp]
	\centering
	\includegraphics[width=\textwidth, keepaspectratio]{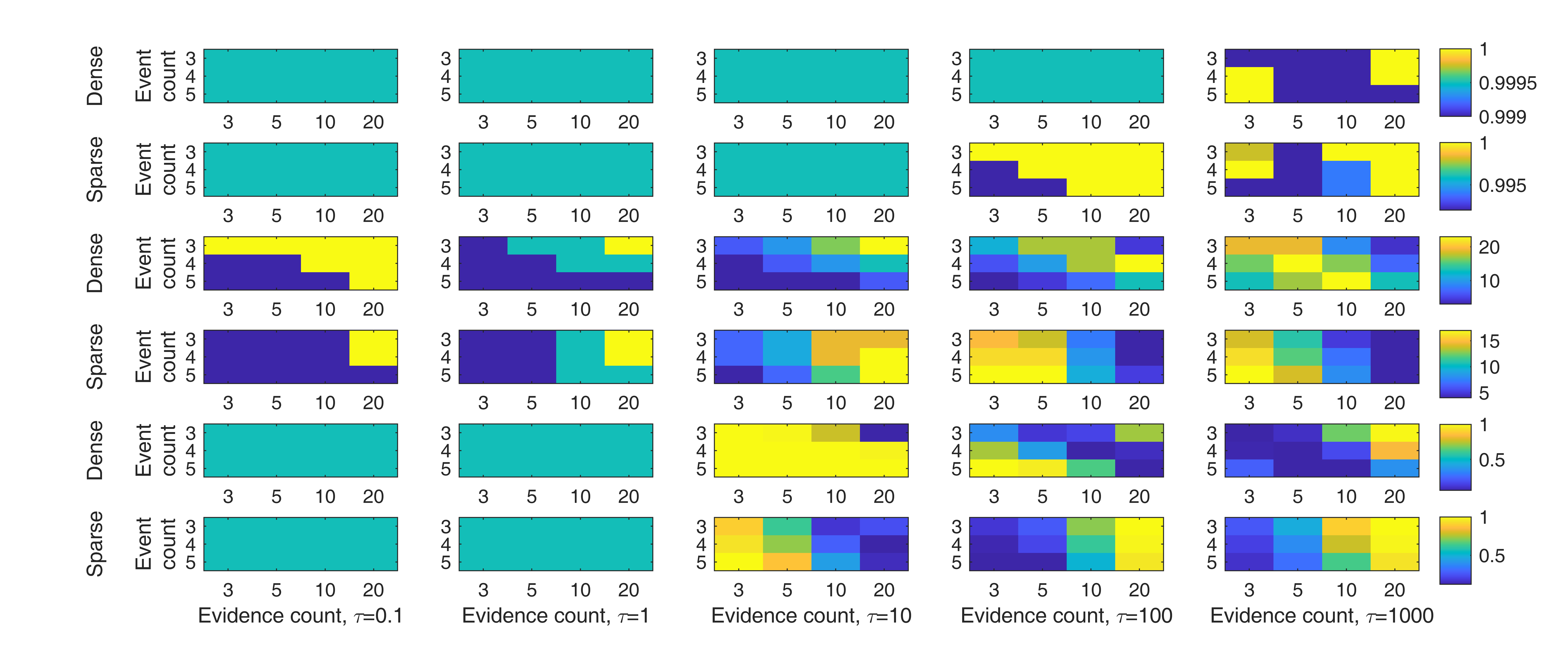}
	\caption{Empirical fixed-point attainment performance of the fast iteration solver under different values of \(\tau\), \(n\), and \(N\). Rows 1--6 from top to bottom: overall attainment rate (dense, sparse), median iteration count among successful trials (dense, sparse), and attainment rate within 10 iterations (dense, sparse). Each row shares a common color bar placed to the right of the fifth column.}
	\label{fig:combined_heatmaps}
\end{sidewaysfigure}
Fig.\ref{fig:combined_heatmaps} reports the first three metrics for the dense and sparse evidence sets. Each row uses a common color scale. The prescribed residual threshold is attained in most trials. Increasing $\tau$ mainly delays attainment, as indicated by the lower attainment rate within 10 iterations and the larger median iteration count. In comparison, $n$, $N$, and evidence sparsity have smaller effects on the attainment speed and overall attainment rate. Thus, within the tested ranges, the distance coefficient primarily affects solution speed rather than eventual attainment.

\begin{figure*}[!t]
	\centering
	\subfloat[Different distance coefficients $\tau$.]{
		\includegraphics[width=0.47\textwidth]
		{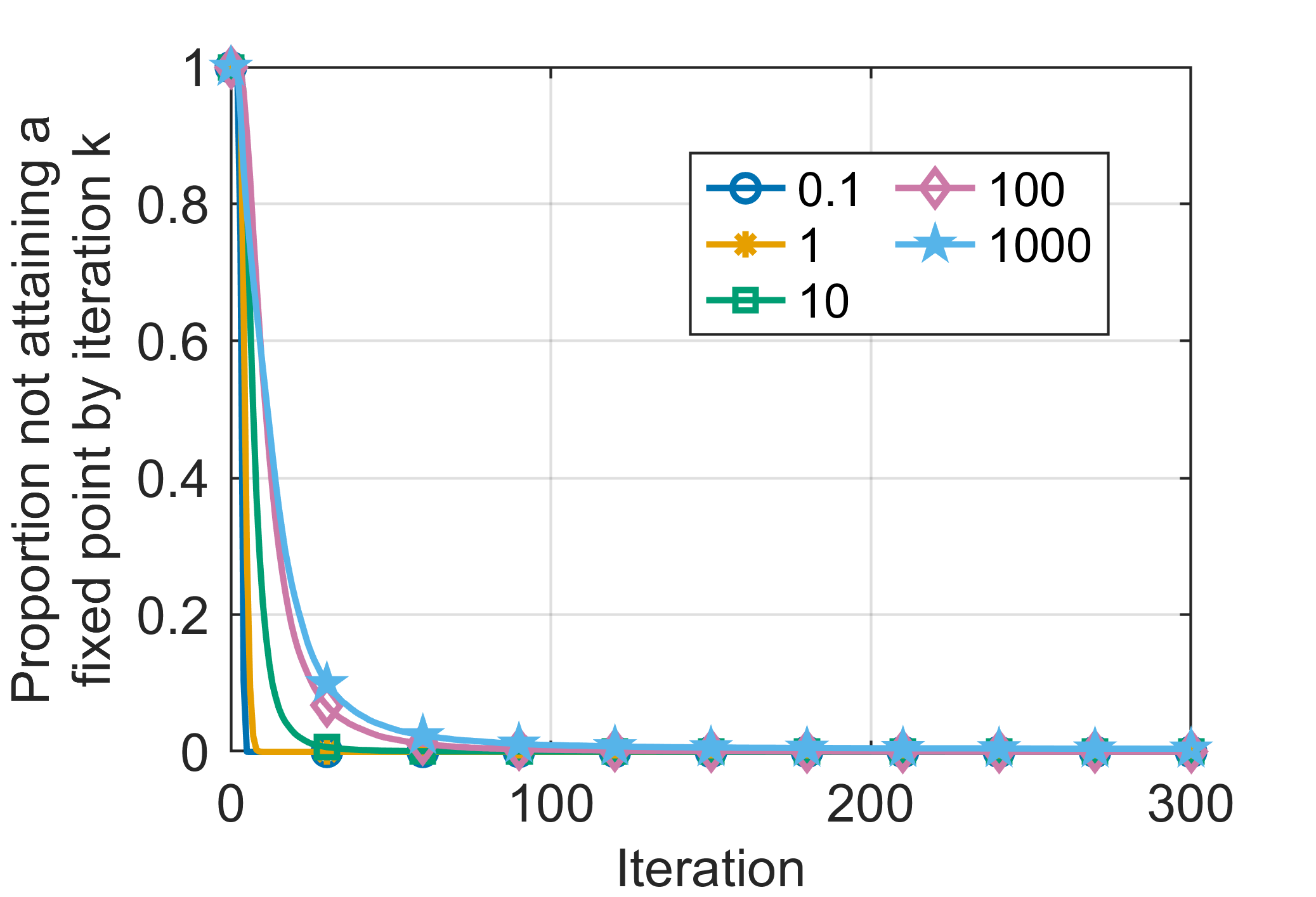}
		\label{fig:unresolved_ratio_tau}
	}
	\hfill
	\subfloat[Different numbers of events $n$.]{
		\includegraphics[width=0.47\textwidth]
		{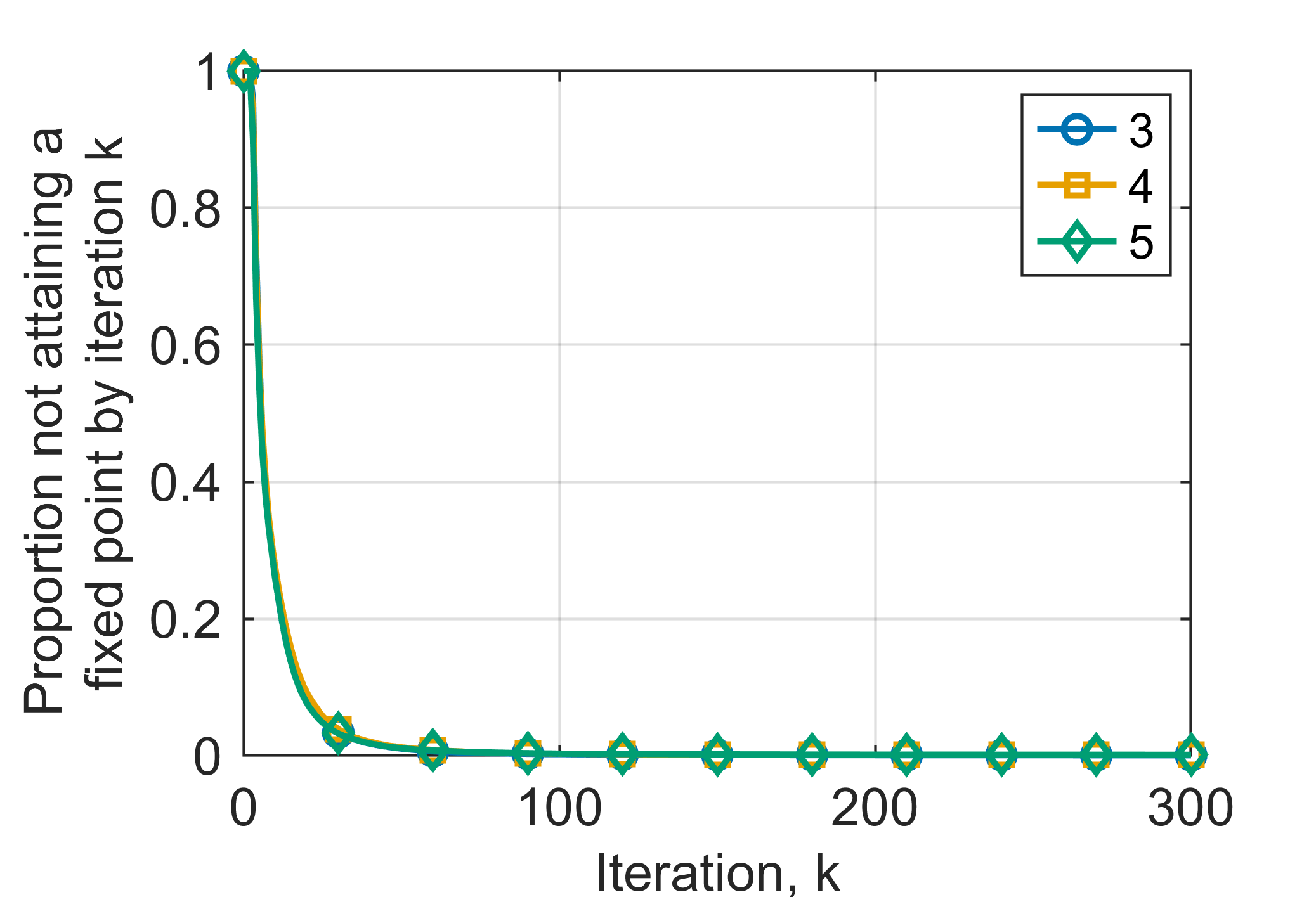}
		\label{fig:unresolved_ratio_n}
	}
	
	\vspace{2mm}
	
	\subfloat[Different numbers of evidences $N$.]{
		\includegraphics[width=0.47\textwidth]
		{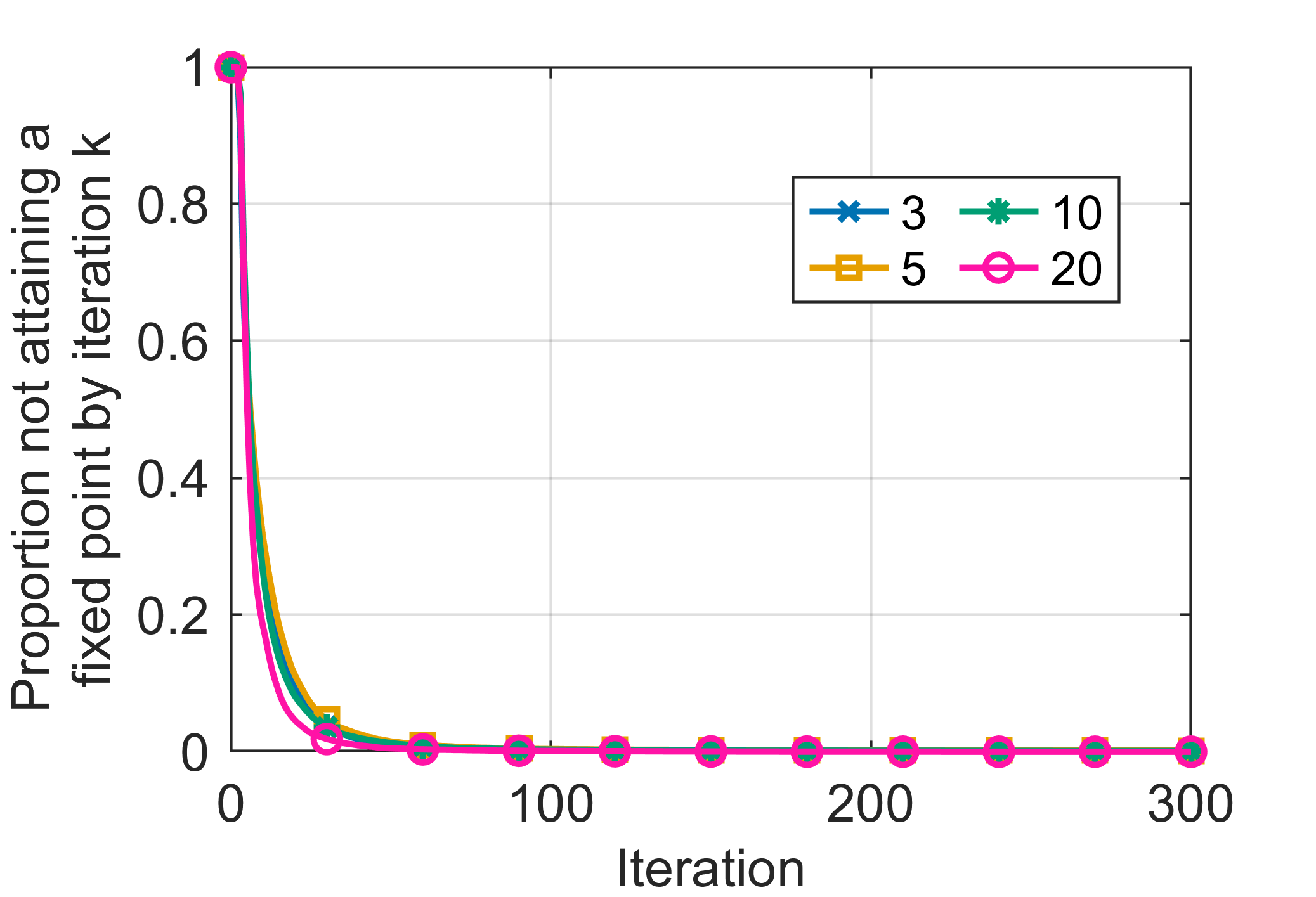}
		\label{fig:unresolved_ratio_N}
	}
	\hfill
	\subfloat[Dense and sparse evidences.]{
		\includegraphics[width=0.47\textwidth]
		{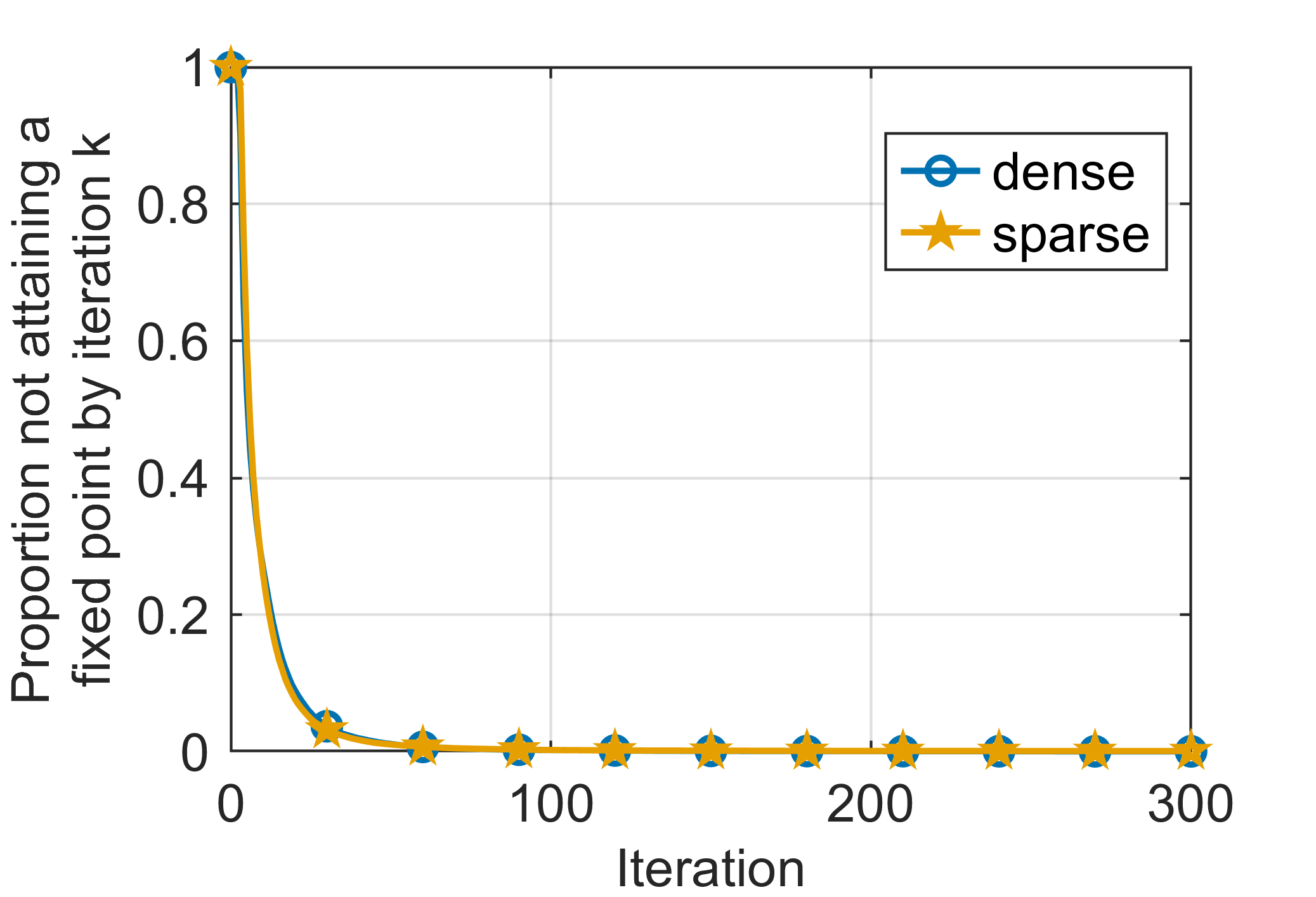}
		\label{fig:unresolved_ratio_mode}
	}
	\caption{Proportions of trials that $r(\boldsymbol{p}^{(k)})>\delta$ after each iteration under different experimental settings.}
	\label{fig:unresolved_ratio}
\end{figure*}
Fig.\ref{fig:unresolved_ratio} further shows that the proportion of unattained trials that have not satisfied the residual threshold after iteration $k$ decreases rapidly during the early iterations. Larger values of $\tau$ produce a slower decrease and a longer tail, consistent with the results in Fig.\ref{fig:combined_heatmaps}. The curves for dense and sparse evidences are close, further indicating that evidence sparsity is not the primary factor affecting the fast fixed-point solver.

These results demonstrate that direct fixed-point iteration with $\epsilon=1$ has a high empirical attainment rate for the randomly generated evidences considered in the experiments and that most successful trials require only a small number of iterations. The results characterize the empirical numerical performance of the fast solver over the tested parameter ranges and do not constitute a theoretical claim of global convergence for arbitrary evidence inputs.

\subsubsection{Periodic Orbit and Relaxed Updates}
\label{subsubsec: Periodic Orbit and Relaxed Updates}
\begin{figure*}[!t]
	\centering
	\includegraphics[width=0.96\textwidth]
	{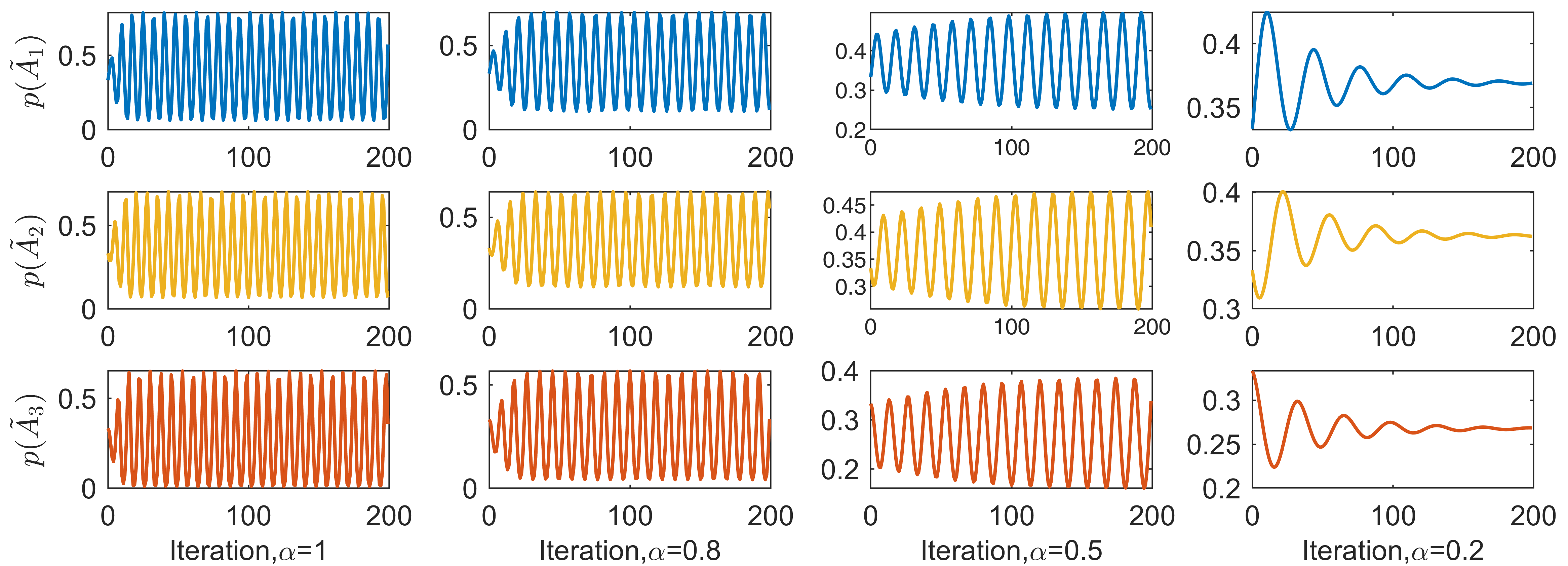}
	\caption{Event-probability trajectories in the periodic-orbit example under different relaxation factors.}
	\label{fig:cycle_probability}
\end{figure*}
\begin{table}[!t]
	\setlength\tabcolsep{0.3mm
		\centering
		\caption{Evidence set producing a periodic orbit under direct fixed-point iteration. ($n=3$)}
		\label{tab:periodic_evidence}
		\renewcommand{\arraystretch}{1.08}
		\begin{tabular}{c|ccccccc}
			\hline
			Evidence & $\{A_1\}$ & $\{A_2\}$ & $\{A_3\}$ & $\{A_1,A_2\}$ & $\{A_1,A_3\}$ & $\{A_2,A_3\}$ & $\Omega$ \\
			\hline
			$\boldsymbol{m}_1$ & $2.800\!\times\!10^{-4}$ & $3.720\!\times\!10^{-7}$ & 0.03729 & 0.9216 & 0.03914 & $5.883\!\times\!10^{-4}$ & $1.053\!\times\!10^{-3}$ \\
			$\boldsymbol{m}_2$ & 0.3993 & 0.01279 & 0.1389 & $1.799\!\times\!10^{-3}$ & 0.4452 & $5.807\!\times\!10^{-6}$ & $2.049\!\times\!10^{-3}$ \\
			$\boldsymbol{m}_3$ & $3.831\!\times\!10^{-8}$ & $9.699\!\times\!10^{-4}$ & 0.05778 & 0.05279 & 0.10199 & 0.7865 & $1.261\!\times\!10^{-11}$ \\
			\hline
		\end{tabular}
	}
\end{table}

\begin{figure}[!t]
	\centering
	\includegraphics[width=0.5\columnwidth]
	{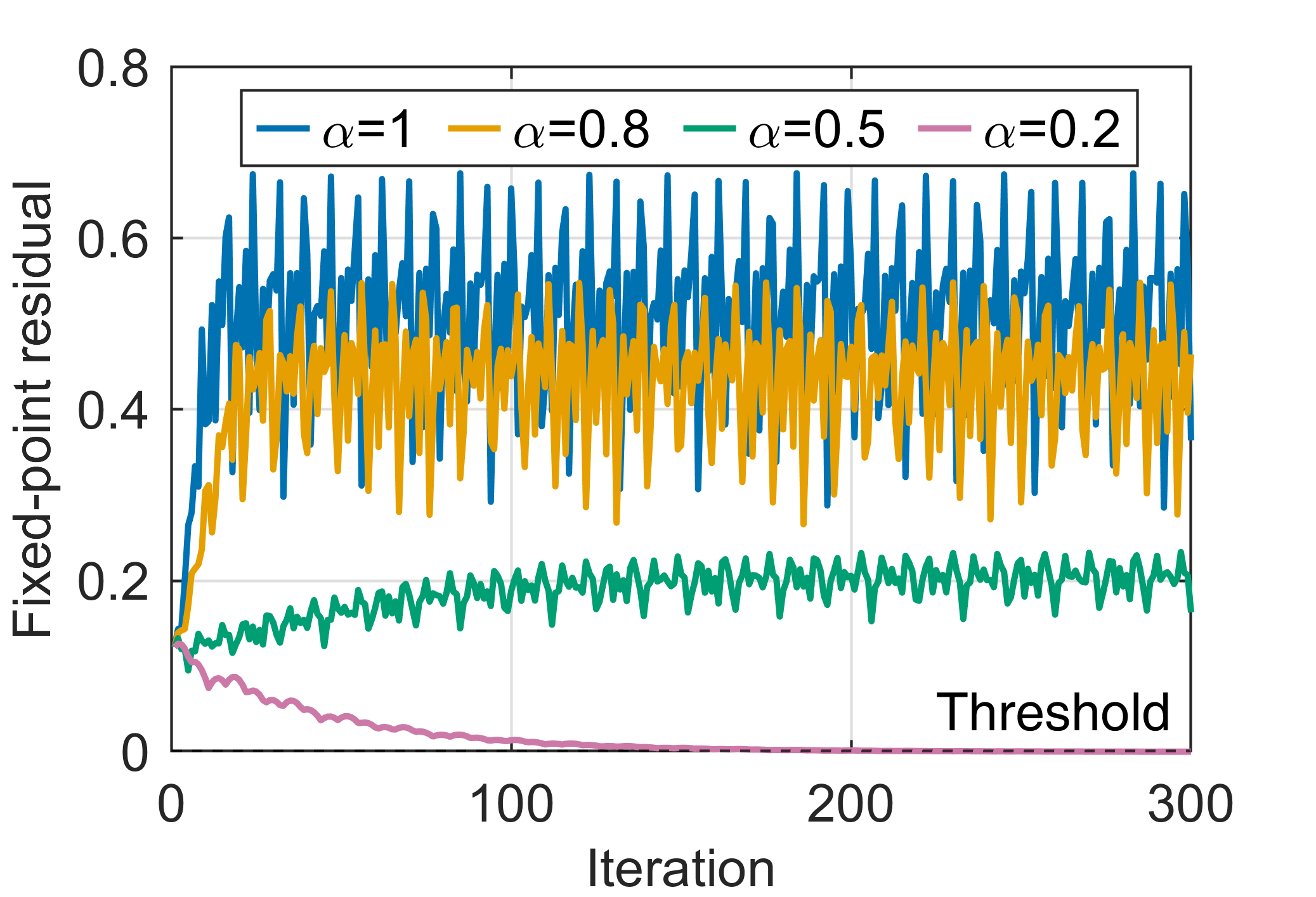}
	\caption{Fixed-point residuals in the periodic-orbit example under different relaxation factors.}
	\label{fig:cycle_residual}
\end{figure}
Although the probability-simplex constraint prevents $\boldsymbol{p}$ from diverging without bound, the iteration
$\boldsymbol{p}^{(k+1)}=\mathcal{F}(\boldsymbol{p}^{(k)})$
may form a periodic orbit for a small number of evidence inputs. For the evidence set in Table~\ref{tab:periodic_evidence}, Fig.\ref{fig:cycle_probability} shows the probability trajectories under different values of $\epsilon$. As $\epsilon$ decreases from 1 to 0.2, the oscillation amplitude is progressively reduced, and the trajectories tend to stabilize at $\epsilon=0.2$. The fixed-point residuals in Fig.\ref{fig:cycle_residual} exhibit the same behavior. Thus, relaxation introduces numerical damping and may suppress a particular periodic orbit, but it does not guarantee fixed-point attainment for every $0<\epsilon<1$.

\subsubsection{Kuhn Simplicial Fixed-Point Search}
\label{subsubsec: Kuhn Simplicial Fixed-Point Search}
\begin{figure}
	\centering
	\includegraphics[width=3.2in]{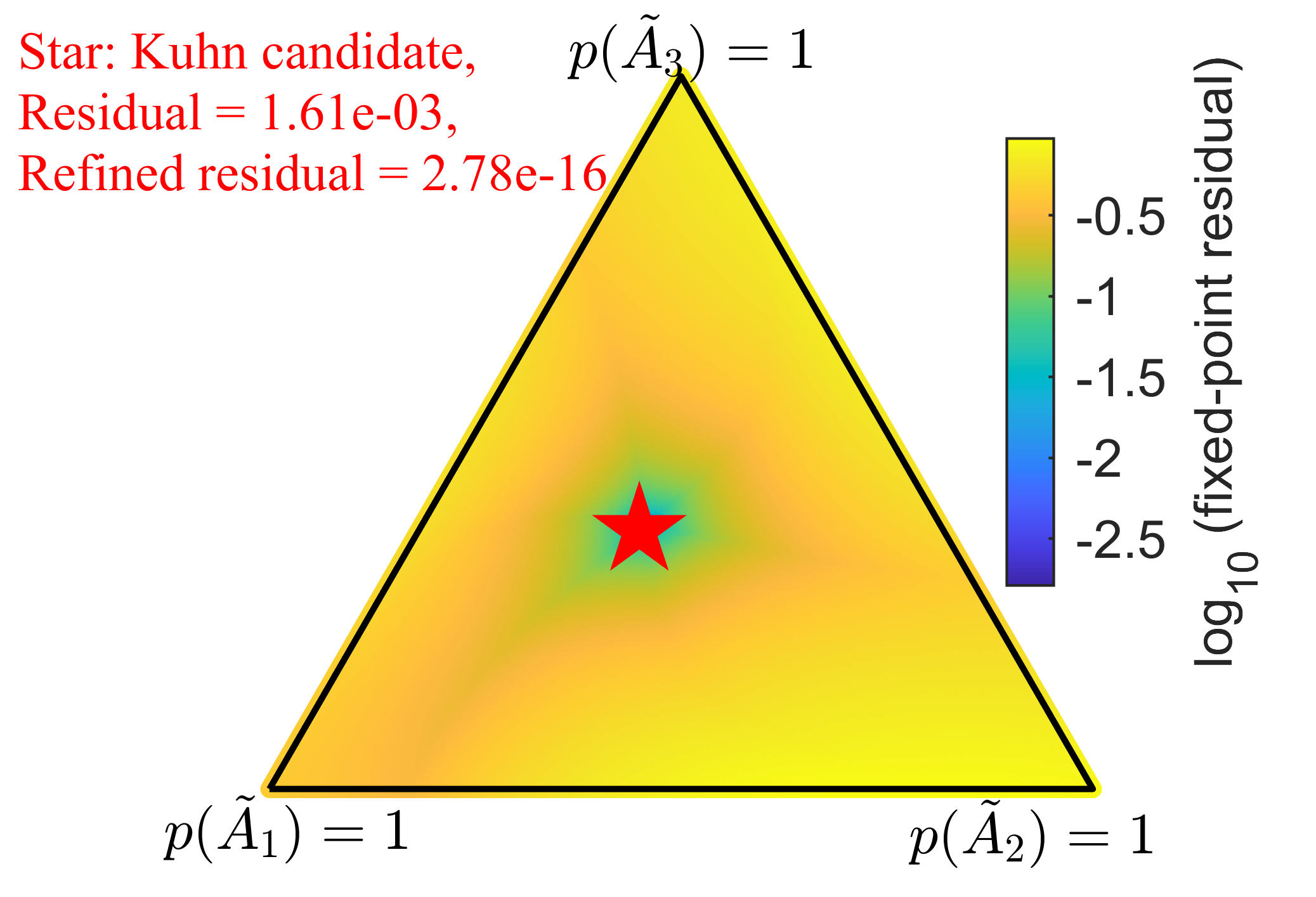}
	\caption{Fixed-point residual distribution over the $\Delta^2$. The colors represent the residuals at the finite-mesh vertices, and the red star denotes the Kuhn candidate.}
	\label{fig:kuhn_simplex}
\end{figure}

This experiment evaluates the ability of the Kuhn simplicial method to locate a fixed-point neighborhood for the evidence set in Table~\ref{tab:periodic_evidence}. Fig.\ref{fig:kuhn_simplex} shows that the low-residual mesh vertices and the candidate obtained from a completely labeled triangle are concentrated in the same region of the probability simplex. Thus, the Kuhn search can locate a fixed-point neighborhood without relying on attainment by direct iteration.

Table~\ref{tab:kuhn_search} reports the candidate residual and computational cost at different values of $L$, the number of equal subdivisions used to discretize each simplex coordinate, with mesh size $h=1/L$. Increasing $L$ from 20 to 160 reduces the candidate residual from $5.98\times10^{-2}$ to $1.61\times10^{-3}$. Further increasing $L$ to 320 does not improve the candidate, although the number of mapping evaluations and runtime continue to increase. The candidate $\boldsymbol{p}=[0.3688,0.2687,0.3625]^{\mathrm{T}}$ obtained at $L=160$ is used to initialize Eq.\eqref{eq:kuhn_local_refinement}. MATLAB \texttt{fsolve} with the Levenberg--Marquardt algorithm obtains 	$\boldsymbol{p}=[0.3693,0.2679,0.3628]^{\mathrm{T}}$
after five iterations, with a fixed-point residual of $2.78\times10^{-16}$, satisfying $\delta=10^{-8}$. The finite-mesh Kuhn search identifies the candidate region, and the stated local numerical implementation removes the remaining mesh-discretization error. The Kuhn method is therefore used as a supplementary solver for periodic iteration, rather than as the default solver, because its computational cost increases rapidly with the mesh resolution.
\begin{table}[!t]
	\centering
	\caption{Kuhn simplicial search at different mesh resolutions}
	\label{tab:kuhn_search}
	\small
	\setlength{\tabcolsep}{2pt}
	\begin{tabular}{c c c c c| c c c c c}
		\hline
		L & h & \makecell{Candidate\\ residual} & \makecell{Map\\ evaluations} & Time (s) &
		L & h & \makecell{Candidate\\ residual} & \makecell{Map\\ evaluations} & Time (s) \\
		\hline
		20 & 0.050000 & $5.98\times10^{-2}$ & 232 & 0.060 &
		160 & 0.006250 & $1.61\times10^{-3}$ & 13042 & 1.891 \\
		40 & 0.025000 & $2.89\times10^{-2}$ & 862 & 0.124 &
		320 & 0.003125 & $1.61\times10^{-3}$ & 51682 & 5.290 \\
		80 & 0.012500 & $1.07\times10^{-2}$ & 3322 & 0.379 &
		-- & -- & -- & -- & -- \\
		\hline
	\end{tabular}
\end{table}
\subsection{Computational Efficiency and Scalability}
\label{subsec: Computational Efficiency and Scalability}
\begin{figure*}[!h]
	\centering
	\begin{minipage}{0.49\linewidth}
		\centering
		\includegraphics[width=3.3in]{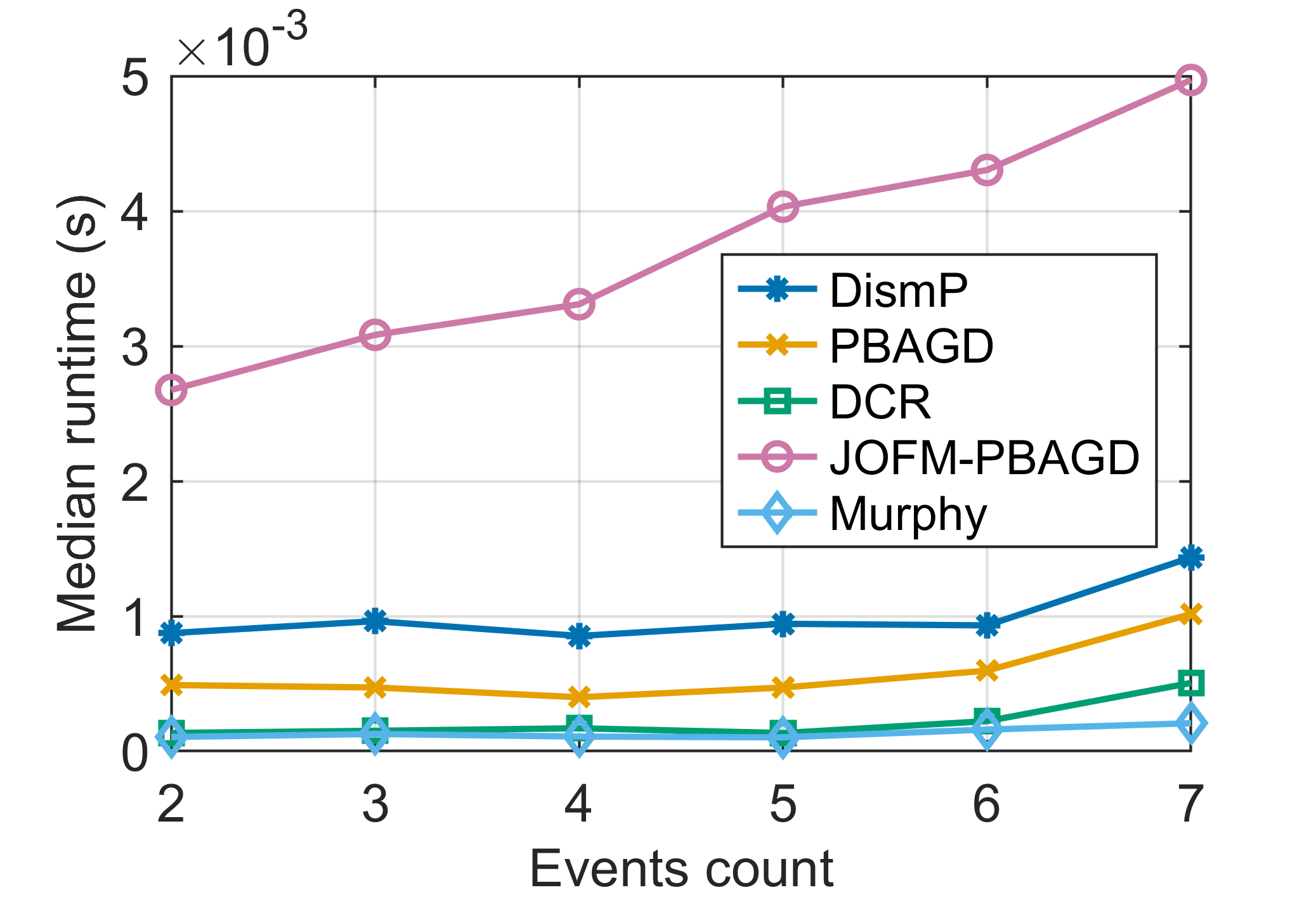}
		\caption{Runtime under a common mapping-evaluation budget with varying numbers of events $n$ and a fixed evidence count of $N=5$.}
		\label{fig:runtime_n}
	\end{minipage}
	\begin{minipage}{0.49\linewidth}
		\centering
		\includegraphics[width=3.3in]{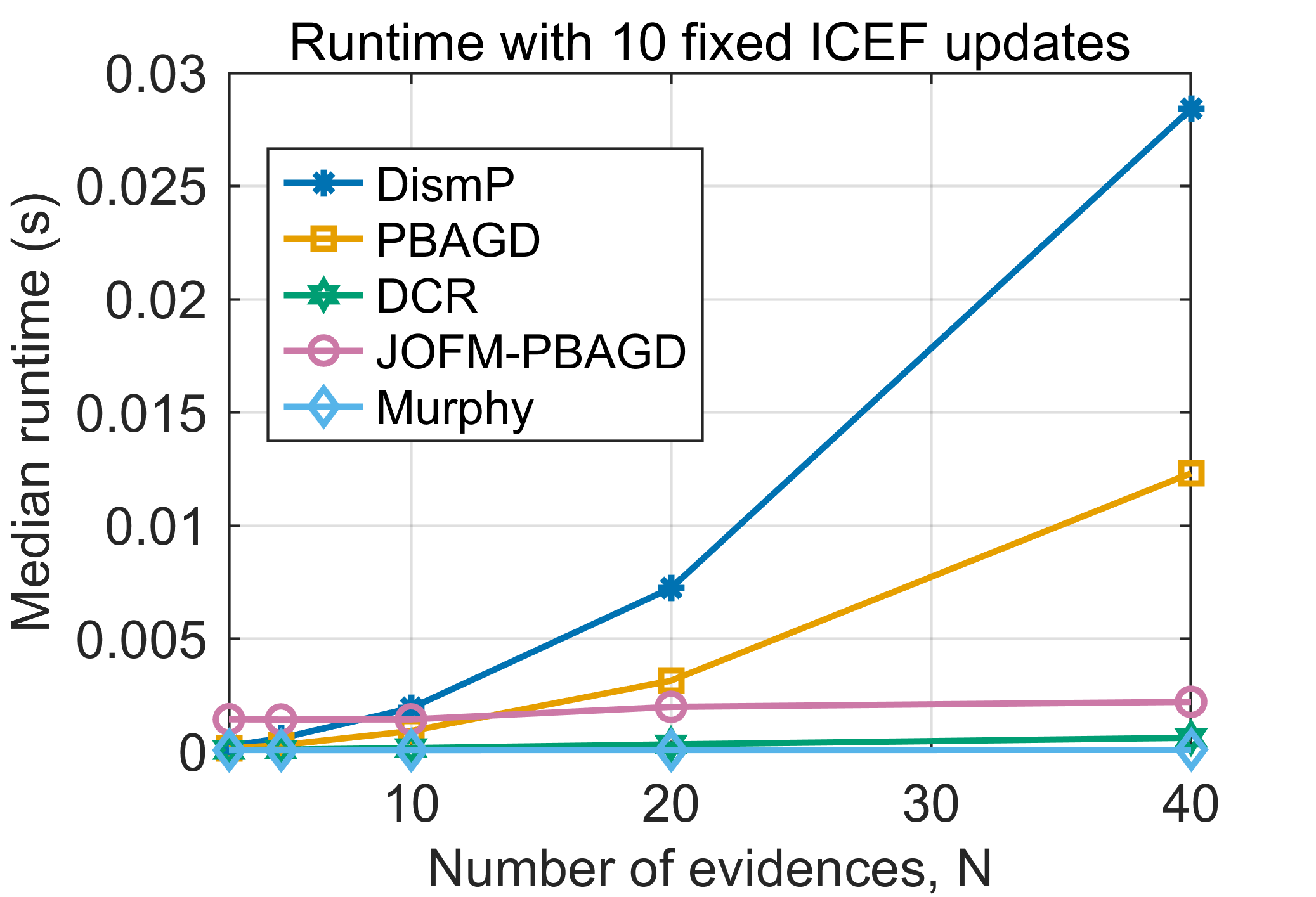}
		\caption{Runtime under a common mapping-evaluation budget with varying numbers of evidences $N$ and a fixed event count of $n=3$.}
		\label{fig:runtime_N}
	\end{minipage}
\end{figure*}

In this subsection, the runtimes of DCR, Murphy, Dismp, PBAGD, and JOFM-PBAGD are compared under varying event and evidence counts. For varying $N$, nested inputs are constructed from the first $N$ evidences of the largest generated evidence set. JOFM-PBAGD performs 10 fixed-point mapping evaluations in each test, providing a common mapping-evaluation budget rather than using the residual-based stopping condition. Each input is evaluated five times, and the median runtime is recorded over 30 randomly generated inputs. As shown in Figs.\ref{fig:runtime_n} and \ref{fig:runtime_N}, the runtimes generally increase with $n$ because the mass-function dimension grows exponentially. JOFM-PBAGD requires more time than the one-shot CEF methods because it repeatedly evaluates the joint-model mapping, but remains on the millisecond scale over the tested range. As $N$ increases, the pairwise comparisons in Dismp and PBAGD cause relatively rapid runtime growth. In JOFM-PBAGD, the event-conditioned credibility matrix is computed only once, so its iterative-stage runtime increases more mildly with $N$.

\begin{figure*}[!h]
	\centering
	\begin{minipage}{0.49\linewidth}
		\centering
		\includegraphics[width=3.3in]{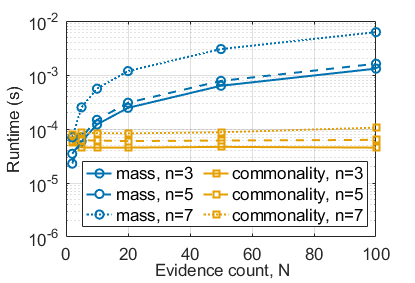}
		\caption{Runtime comparison between sequential DCR in the mass-function domain and exponentiation in the commonality-function domain.}
		\label{fig:mass_commonality_time}
	\end{minipage}
	\begin{minipage}{0.49\linewidth}
		\centering
		\includegraphics[width=3.3in]{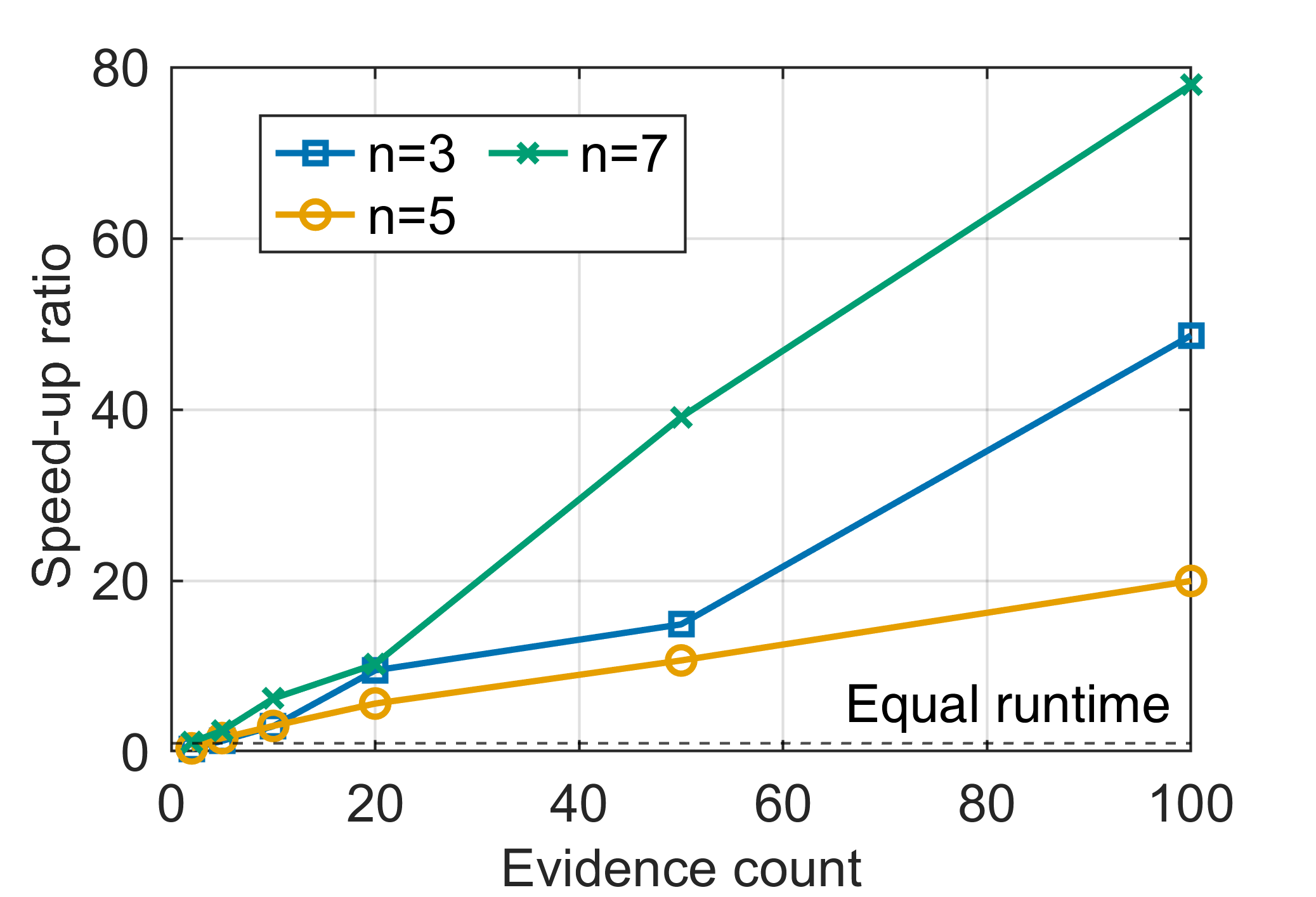}
		\caption{Speed-up ratios achieved by the commonality-function implementation for different numbers of events and evidences.}
		\label{fig:commonality_speedup}
	\end{minipage}
\end{figure*}

The acceleration provided by Eq.\eqref{eq:commonality_power} is further evaluated by comparing sequential DCR in the mass-function domain with elementwise exponentiation in the commonality-function domain. The experiment considers $n\in\{3,5,7\}$ and $N\in\{2,5,10,20,50,100\}$. Evidences with nested or intersecting focal elements are used to avoid total self-conflict. Figs.\ref{fig:mass_commonality_time} and \ref{fig:commonality_speedup} show that sequential DCR becomes increasingly expensive as $N$ grows, whereas the commonality implementation mainly requires one forward transform, elementwise exponentiation, and one inverse transform. Consequently, the speed-up ratio $S=T_{\mathrm{mass}}/T_{\mathrm{commonality}}$ increases with $N$, supporting commonality-function exponentiation for repeatedly combining identical average evidences.

\subsection{Applicability to Different Evidence Dissimilarity Measures}
\label{subsec: Applicability to Different Evidence Dissimilarity Measures}

Event-conditioned credibility is calculated from the EDM between an evidence and a candidate-event evidence, but the JOFM does not prescribe a particular EDMF. This experiment uses RB, BJS, Dismp, PBLBJS, PBAGD, and RFBD to construct event-conditioned credibility and fuses the evidences in Table~\ref{tab6:evidenceExample}.
\begin{table}[!h]
	\setlength\tabcolsep{1.5mm}   
	\centering
	\renewcommand{\arraystretch}{1}
	\caption{Multisensors evidence \cite{anevidential2018xia}.}
	\label{tab6:evidenceExample}
	\begin{tabular}{cccccc|cccccc}   
		\toprule[1.25pt]
		& $\boldsymbol{m}_1$ & $\boldsymbol{m}_2$ & $\boldsymbol{m}_3$ & $\boldsymbol{m}_4$ & $\boldsymbol{m}_5$ &
		& $\boldsymbol{m}_1$ & $\boldsymbol{m}_2$ & $\boldsymbol{m}_3$ & $\boldsymbol{m}_4$ & $\boldsymbol{m}_5$ \\
		\midrule[0.5pt]
		$\{\tilde{A}_1\}$ & 0.50 & 0.00 & 0.55 & 0.55 & 0.60 &
		$\{\tilde{A}_2\}$ & 0.20 & 0.90 & 0.10 & 0.10 & 0.10 \\
		$\{\tilde{A}_3\}$ & 0.30 & 0.10 & 0.00 & 0.00 & 0.00 &
		$\{\tilde{A}_1,\tilde{A}_3\}$ & 0.00 & 0.00 & 0.35 & 0.35 & 0.30 \\
		\bottomrule[1.25pt]
	\end{tabular}
\end{table}

\begin{table}[!h]
	\setlength\tabcolsep{2mm}
	\centering
	\renewcommand{\arraystretch}{1}
	\caption{Evidence credibility under six EDMFs. }
	\label{tab:credibilityRes_Xiao}
	\begin{tabular}{cccccc}
		\toprule [1.25pt]
		Method&$\boldsymbol{m}_1$&$\boldsymbol{m}_2$&$\boldsymbol{m}_3$&$\boldsymbol{m}_4$&$\boldsymbol{m}_5$ \\ 
		\midrule[0.5pt]
		JOFM-RB		&	0.1734		&0.0979&	0.2403	   &	0.2403	   &\textbf{0.2480}	\\
		JOFM-BJS	&	0.2047		&0.0652&	0.2316	   &	0.2316	   &\textbf{0.2669}	\\
		JOFM-Dismp	&	0.1407		&0.0714&	0.2542	   &	0.2542	   &\textbf{0.2795}	\\
		JOFM-PBLBJS	&	0.1320		&0.0418&	0.2655	   &	0.2655	   &\textbf{0.2951}	\\
		JOFM-PBAGD	&	0.0831		&0.0170&	0.2789	   &	0.2789	   &\textbf{0.3422}	\\
		JOFM-RFBD	&	0.1727		&0.0975&	0.2419	   &	0.2419	   &\textbf{0.2460}	\\
		\bottomrule [1.25pt]
	\end{tabular}
\end{table}

As shown in Table~\ref{tab:credibilityRes_Xiao}, all six JOFM variants assign the highest credibility to $\boldsymbol{m}_5$ and the lowest one to $\boldsymbol{m}_2$. The $\boldsymbol{m}_5$ provides the strongest support for the ground truth $\tilde{A}_1$, whereas $\boldsymbol{m}_2$ mainly supports the incorrect event $\tilde{A}_2$. It indicates that different EDMFs, if incorporated into the proposed JOFM, will produce credibility rankings that reflect the relative support of the evidences for the candidate decision events.

\begin{table}[!h]
	\setlength\tabcolsep{0.3mm
		\centering
		\renewcommand{\arraystretch}{1}
		\caption{Fusion results for evidence in Table \ref{tab6:evidenceExample}.}
		\label{tab7:Fus_Res_tab6}
		\begin{tabular}{ccccc|cccccc}
			\toprule [1.25pt]
			Method&$\{\tilde A_1\}$&$\{\tilde A_2\}$&$\{\tilde A_3\}$&$\{\tilde A_1,\tilde A_3\}$
			&Method&$\{\tilde A_1\}$&$\{\tilde A_2\}$&$\{\tilde A_3\}$&$\{\tilde A_1,\tilde A_3\}$ \\
			\midrule[0.5pt]
			JOFM-RB		&	0.9873		&	0.0013	&	0.0078	&	0.0037	&
			JOFM-BJS	&	0.9881		&	0.0006	&	0.0080	&	0.0033	\\
			JOFM-Dismp	&	0.9894		&	0.0005	&	0.0058	&	0.0042	&
			JOFM-PBLBJS	&	0.9905		&	0.0002	&	0.0049	&	0.0045	\\
			JOFM-PBAGD	&\textbf{0.9919}&	0.0001	&	0.0028	&	0.0052	&
			JOFM-RFBD	&	0.9873		&	0.0012	&	0.0078	&	0.0037	\\
			\bottomrule [1.25pt]
	\end{tabular}}
\end{table}
Table~\ref{tab7:Fus_Res_tab6} presents the corresponding fusion results. All six JOFM variants correctly identify $\tilde{A}_1$ and assign masses ranging from 0.9873 to 0.9919 to this event, and JOFM-PBAGD provides the highest support in this example, which shows that the JOFM does not depend on a particular EDMF. Different measures nevertheless produce different relative event-conditioned credibility values and consequently affect the overall credibility and fusion result.

\section{Conclusion}
\label{sec: conclusion}
This paper developed a decision-oriented joint optimization model for evidence fusion to reduce the underestimation of critical evidence caused by credibility calculation based only on inter-evidence comparisons. Event-conditioned credibility relates evidence credibility to candidate-event hypotheses, while PBAGD measures the difference between evidence and candidate-event evidence. The coupled credibility--fusion--decision process was formulated as a continuous self-mapping on the candidate-event probability simplex, establishing the existence of at least one fixed point. Direct fixed-point iteration attained the prescribed residual threshold in most Monte Carlo trials, although a constructed example showed that it may enter a periodic orbit. For this case, the Kuhn simplicial search combined with local refinement successfully located a numerical fixed point. Numerical examples further showed that the proposed method increased the credibility of evidence strongly supporting the ground truth, suppressed anomalous evidence, and improved the fused support for the correct event.

%

\section*{Acknowledgements}
This work is supported by the National Natural Science Foundation of China under Grant 62573357.


\bibliography{Reference}
\end{document}